%% file: neurips_2026.tex
\documentclass{article}

    \PassOptionsToPackage{numbers, compress}{natbib}
 \usepackage[preprint]{neurips_2026}

\usepackage[utf8]{inputenc} 
\usepackage[T1]{fontenc}    
\usepackage{hyperref}       
\usepackage{url}            
\usepackage{booktabs}       
\usepackage{amsfonts}       
\usepackage{nicefrac}       
\usepackage{microtype}      
\usepackage{xcolor}         

\usepackage{graphicx}
\usepackage{subcaption}

\usepackage{amsmath}
\usepackage{amssymb}
\usepackage{mathtools}
\usepackage{amsthm}

\input{shortcuts}

\usepackage[capitalize,noabbrev]{cleveref}

\theoremstyle{plain}
\newtheorem{theorem}{Theorem}[section]
\newtheorem{proposition}[theorem]{Proposition}

\theoremstyle{definition}

\theoremstyle{remark}
\newtheorem{remark}[theorem]{Remark}

\title{T-FIX: Text-Based Explanations with Features Interpretable to Experts}

\usepackage{pifont}
\usepackage{wasysym}

\usepackage{bbding}

\newcommand{\penncis}{\textsuperscript{\ensuremath{\spadesuit}}}
\newcommand{\utaustin}{\textsuperscript{\ensuremath{\clubsuit}}}
\newcommand{\pennphysics}{\textsuperscript{$\bigstar$}}
\newcommand{\flatiron}{$^{\text{\raisebox{-0.45ex}{\scalebox{0.9}{\FiveStarOpen}}}}$}
\newcommand{\pennmedsurg}{\textsuperscript{\ensuremath{\dagger}}}
\newcommand{\pennmedsepsis}{\textsuperscript{\ensuremath{\maltese}}}
\newcommand{\pennmedcardiology}{\textsuperscript{$\heartsuit$}}
\newcommand{\toronto}{\textsuperscript{\ensuremath{\ddagger}}}
\newcommand{\uhn}{\textsuperscript{\ensuremath{\diamondsuit}}}

\author{%
Shreya Havaldar\penncis\,\thanks{\raisebox{-0.2ex}{\includegraphics[height=1.9ex]{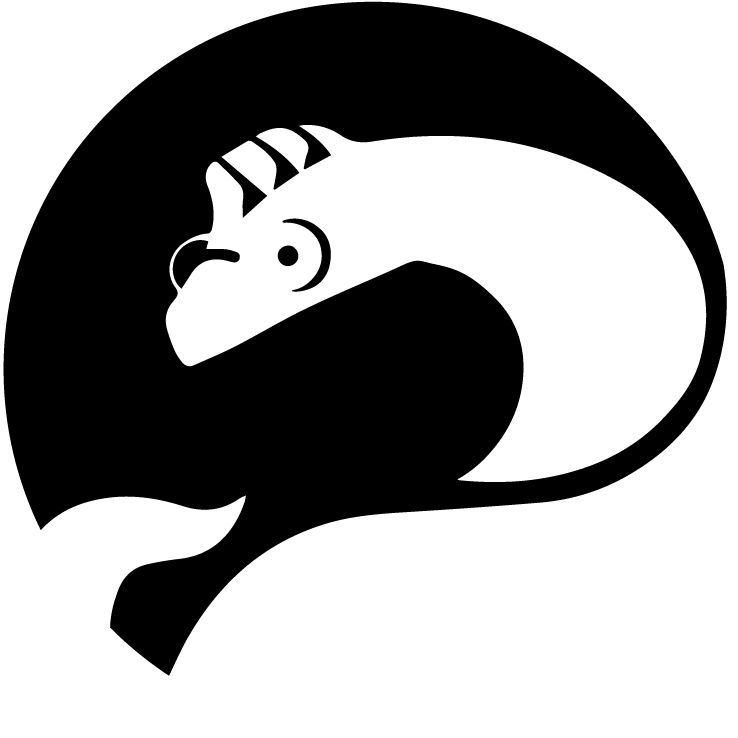}} Equal contribution. Code is available at \ourrepo.}\quad
Weiqiu You\penncis$^*$\,\quad
Chaehyeon Kim\penncis\quad
Anton Xue\utaustin\quad
Helen Jin\penncis\\
\textbf{Marco Gatti}\pennphysics\quad
\textbf{Bhuvnesh Jain}\pennphysics\quad
\textbf{Helen Qu}\flatiron\quad
\textbf{Amin Madani}\toronto\uhn\quad
\textbf{Daniel A. Hashimoto}\pennmedsurg\penncis\\
\textbf{Gary E. Weissman}\pennmedsepsis\quad
\textbf{Rajat Deo}\pennmedcardiology\quad
\textbf{Sameed Khatana}\pennmedcardiology\quad
\textbf{Lyle Ungar}\penncis\quad
\textbf{Eric Wong}\penncis\\[2ex]
{\small
\penncis Department of Computer and Information Science, University of Pennsylvania;}\\
{\small
\utaustin Department of Computer Science, University of Texas at Austin;}\\
{\small
\pennphysics Department of Physics and Astronomy, University of Pennsylvania;
\flatiron Flatiron Institute;}\\
{\small
\pennmedsurg Department of Surgery, Perelman School of Medicine, University of Pennsylvania;}\\
{\small
\pennmedsepsis Division of Pulmonary, Allergy, and Critical Care, Perelman School of Medicine, University of Pennsylvania;}\\
{\small
\pennmedcardiology Division of Cardiovascular Medicine, Perelman School of Medicine, University of Pennsylvania;}\\
{\small
\toronto Department of Surgery, University of Toronto;
\uhn University Health Network.}\\[1ex]
{\small \texttt{\{shreyah,weiqiuy\}@seas.upenn.edu}}
}

\begin{document}

\maketitle

\input{sections/abstract}

\input{sections/introduction}

\input{sections/method}

\input{sections/pipeline_new}

\input{sections/pipeline_validation}

\input{sections/experiments}

\input{sections/discussion}

\input{sections/related}
\input{sections/conclusion}

\input{sections/acknowledgement}


\bibliographystyle{plainnat}
\bibliography{custom}


\appendix

\input{sections/appendix}
\end{document}

%% file: shortcuts.tex
\usepackage[utf8]{inputenc} 
\usepackage[T1]{fontenc}    
\usepackage{url}            
\usepackage{booktabs}       
\usepackage{amsfonts}       
\usepackage{nicefrac}       
\usepackage{microtype}      
\usepackage{xcolor}         
\usepackage{bbm}

\usepackage{graphicx}

\usepackage{amsmath}
\usepackage{amsthm}
\usepackage{amssymb}

\usepackage{makecell}
\usepackage{tabularx}

\usepackage[capitalize,noabbrev]{cleveref}

\definecolor{prussian}{HTML}{006fd4}

\definecolor{skyblue}{HTML}{448EE4}

\usepackage[table]{xcolor}
\usepackage{bm}
\newcommand{\best}[1]{\bm{#1}}
\newcommand{\secondbest}[1]{\underline{#1}}
\newcommand{\bestoverall}[1]{\cellcolor{gray!15}\textbf{#1}}

\usepackage{xspace}
\newcommand{\dataset}{\textsc{T-FIX}\xspace}
\newcommand{\score}{\mathsf{T\textrm{-}FIXscore}}

\newcommand{\Om}{$\Omega_m$}
\newcommand{\seight}{$\sigma_{8}$}
\newcommand{\ourrepo}{{\small\url{https://github.com/BrachioLab/FIX-2}}}
\newcommand{\groupalign}{\mathsf{ExpertAlign}\xspace}
\newcommand{\claimalign}{\mathsf{ClaimAlign}\xspace}
\newcommand{\featurealign}{\mathsf{FeatureAlign}\xspace}
\newcommand{\featuregroupalign}{\mathsf{ExpertAlign}\xspace}
\newcommand{\fixscore}{\mathsf{FIXscore}\xspace}

\usepackage{thm-restate}

\input{urls}

\usepackage{mathtools}

\definecolor{myblue}{HTML}{0000ff}

\usepackage{amsmath}

\usepackage{array, multicol, multirow}
\usepackage{enumitem}

\usepackage{tcolorbox}
\tcbuselibrary{listings} 
\newtcblisting{promptbox}{
  listing only,
  listing options={
    basicstyle=\ttfamily\scriptsize,
    breaklines=true,
    prebreak=\mbox{\tiny\ensuremath\hookleftarrow},
  },
  colback=gray!10,
  colframe=violet,
  left=2mm, right=2mm, top=1mm, bottom=1mm,
  boxrule=0.5pt,
  arc=1pt,
  outer arc=1pt,
  title=Prompt,
  fonttitle=\bfseries,
  sharp corners,
}

\usepackage{tcolorbox}
\tcbuselibrary{listings} 
\newtcblisting{baselinepromptbox}{
  listing only,
  listing options={
    basicstyle=\ttfamily\scriptsize,
    breaklines=true,
    prebreak=\mbox{\tiny\ensuremath\hookleftarrow},
  },
  colback=gray!10,
  colframe=teal,
  left=2mm, right=2mm, top=1mm, bottom=1mm,
  boxrule=0.5pt,
  arc=1pt,
  outer arc=1pt,
  title=Prompt,
  fonttitle=\bfseries,
  sharp corners,
}

%% file: sections/abstract.tex
\begin{abstract}
As LLMs are deployed in knowledge-intensive settings (e.g., surgery, astronomy, therapy), users are often domain experts who expect not just answers, but explanations that mirror professional reasoning. 
Yet evaluating whether an LLM ``thinks like an expert'' remains difficult: existing approaches rely on per-example expert annotation, making them costly, hard to scale, and tied to a single notion of correct reasoning within each domain.
To address this gap, we introduce \dataset, a unified evaluation framework that operationalizes \textit{expert alignment} as a desired attribute of LLM-generated explanations.
\dataset spans seven scientific tasks across three domains, with each task evaluated against expert-defined criteria that capture domain-grounded reasoning rather than generic explanation quality. Our framework enables automatic, personalizable evaluation of expert alignment that generalizes to unseen explanations without ongoing expert involvement.
\end{abstract}


%% file: sections/introduction.tex
\section{Introduction}
\label{sec:introduction}

LLMs are increasingly used for domain-specific tasks requiring specialized knowledge, with adoption explored in high-stakes environments like operating rooms, astronomical observatories, and therapy clinics \cite{pressman2024clinical, fouesneau2024rolelargelanguagemodels, stade2024large}. For LLMs to be trustworthy in these settings, users require not only correct answers but also good explanations \cite{rudin2019stop, pedreschi2019meaningful}. However, what constitutes a ``good explanation'' depends on the target audience \cite{ribera2019can, sokol2020one}. Standard evaluations of LLM explanations focus on task-agnostic quality dimensions such as plausibility, faithfulness, verbosity, or soundness \cite{zhou2021evaluating, agarwal2024faithfulness, parcalabescu2023measuring, lee-hockenmaier-2025-evaluating, you-etal-2025-probabilistic}. While necessary, these dimensions are insufficient in knowledge-intensive domains, where the primary users --- clinicians, astrophysicists, psychologists --- need reasoning they can verify against domain-validated principles \cite{wang2021explanations}. An LLM may produce a correct answer with a coherent explanation that nonetheless rests on features irrelevant to expert reasoning. In high-stakes settings, such unverified reasoning is itself a safety and trustworthiness concern.

We call this missing dimension \textbf{expert alignment}: the degree to which an explanation's stated claims reflect the concepts, evidence, and inference patterns that domain experts consider relevant for a task. \citet{actionable} advocate for actionable interpretability, where alignment to human intent and meaningfulness to experts are key requirements. This builds on a broader consensus that explanations should match human reasoning \citep{jacovi-2021-aligning, miller2019explanation, doshi-velez_towards_2017} and demonstrably influence real-world decisions \citep{spillner2025can}. Yet current explainability methods remain unreliable in clinical practice \citep{ghassemi2021false}, motivating metrics that directly measure expert-relevant content.

However, measuring expert alignment is difficult. Historically, this has been done by (1) small-scale expert studies, which require substantial manual labor and are not useful for large-scale benchmarking ~\citep{kutbi2025evaluating,hyk2025queriescriteriaunderstandingastronomers}, and (2) larger domain-specific benchmarks, which scale, but use fixed per-example rubrics, limiting generalization to unseen explanations~\citep{li2025astronomical,li2025counselbenchlargescaleexpertevaluation,xu2025mentalchat16k,yang2026healthscorescalablerubricsimproving}. This difficulty is compounded by the heterogeneity of expert reasoning across domains --- cosmology relies on physical structure, psychology on behavioral cues, medicine on safety-critical clinical reasoning~\citep{hyk2025queriescriteriaunderstandingastronomers,ghassemi2021false,miller2019explanation} --- yet building disjoint evaluations for each field is impractical~\citep{doshi-velez_towards_2017,arya2019one}. We require a unified framework that is both \textit{automatic}, enabling large-scale benchmarking, and \textit{generalizable} to unseen explanations.

To fill this gap, we introduce the \textbf{\dataset benchmark}: a suite of seven datasets spanning three scientific domains with an accompanying expert-designed evaluation framework that is both \textit{automatic} and \textit{generalizable} (Figure~\ref{fig:criteria}). Developed in close collaboration with domain experts, \dataset provides, to our knowledge, the first scalable, unified measure of expert alignment across diverse real-world domains and modalities. In summary, our contributions are as follows:
\begin{itemize}
\vspace{-0.1cm}
\itemsep 0.2em
    \item We develop \dataset, the first automatic and generalizable benchmark for expert alignment in LLM-generated explanations. Unifying seven scientific tasks across cosmology, psychology, and medicine, \dataset front-loads expert effort, enabling automatic scoring of unseen explanations without experts needing to annotate additional data points.

     \item The core of our approach, $\score$, is an expert alignment metric for free-form text that satisfies duplication invariance, robustness to ungrounded content, and is sensitive to compounded expert reasoning.

    \item We validate \dataset through human annotation and expert interviews, confirming that our automatic pipeline reliably captures expert judgments about aligned and misaligned reasoning. This enables fully automatic deployment without additional expert effort.

    \item We benchmark six frontier LLMs under four prompting strategies and find that high task performance does not imply expert-aligned explanations, identifying expert alignment as a distinct open challenge for current LLMs.
\end{itemize}

\begin{figure*}[t]
    \centering
    \includegraphics[width=0.9\linewidth]{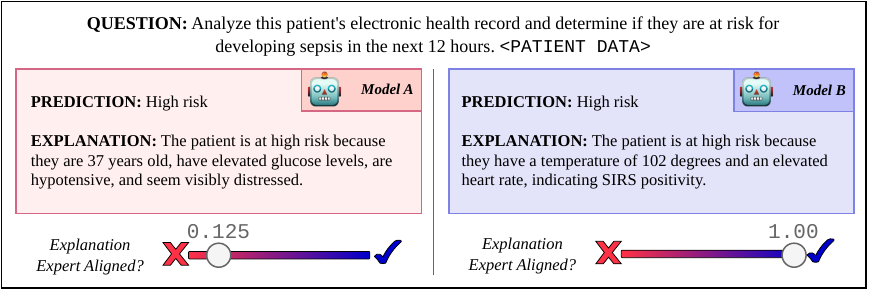}
\caption{Two models both correctly predict high sepsis risk, but their explanations differ in \textbf{expert alignment}. Model~A cites features clinicians would not prioritize (e.g., age of 37, visible distress), scoring low. Model~B references clinically grounded indicators (temperature of 102\textdegree F, elevated heart rate, SIRS positivity), scoring high. In scientific and clinical settings, explanations misaligned with expert reasoning may rely on spurious features, undermining trust even when predictions are correct.}
    \label{fig:abstract-example}
\end{figure*}

%% file: sections/method.tex
\section{Benchmarking Expert Alignment for Text-Based Explanations}
\label{sec:method}

Evaluating whether an explanation is \textit{expert-aligned} --- whether it reflects the concepts, evidence, and inference patterns that domain experts consider relevant --- requires benchmarks that go beyond task-agnostic quality dimensions like plausibility or faithfulness~\cite{actionable,doshi-velez_towards_2017,miller2019explanation}.

Unfortunately, many current methods for measuring expert alignment of LLM outputs across scientific fields \textbf{\textit{are not automatic, and thus hard to scale}}. These small-scale expert studies capture domain reasoning but require substantial manual labor: \citet{kutbi2025evaluating} annotate only 18 sepsis-diagnosis explanations with 7 physicians, and \citet{hyk2025queriescriteriaunderstandingastronomers} report four weeks of interviews with 11 astronomers. Conversely, larger benchmarks for domain-specific LLM evaluation rarely assess expert alignment in a substantive way. Experts appear in validation or annotation studies (e.g., Astro-QA~\citep{li2025astronomical}, CounselBench~\citep{li2025counselbenchlargescaleexpertevaluation}, Mentalchat16k~\citep{xu2025mentalchat16k}), rather than being involved in the benchmark generation process. In medicine, newer efforts introduce per-example, domain-specific rubrics (e.g., Health-SCORE from \citet{yang2026healthscorescalablerubricsimproving} and OpenAI's HealthBench), but \textbf{\textit{generalization to unseen examples is not straightforward, as expert annotations only apply to examples in the benchmark.}}

\subsection{FIX: Automatic and Generalizable Expert Alignment for Feature Groups}
However, scalable and generalizable evaluations exist for non-text-based explanations --- \citet{jin2024fix} propose the FIX benchmark, limited to \emph{feature-group} explanations. Given a task with input $x\in \mathbb{R}^d$, output $y$, and a set of feature groups $\hat{G}$ (binary masks over input features), FIX scores each feature by averaging the alignment of all groups containing it, then averages across features:
\begin{equation}
    \fixscore(\hat{G}, x) =  \frac{1}{d} \sum_{i=1}^{d} \featurealign(i, \hat{G}, x) = \frac{1}{d} \sum_{i=1}^{d} \frac{1}{|\hat{G}[i]|} \sum_{\hat{g} \in \hat{G}[i]} \featuregroupalign(\hat{g}, x),
\end{equation}
where $\hat{G}[i] = \{\hat{g} \in \hat{G} : i \in \hat{g}\}$ is the set of groups containing feature $i$, $\featuregroupalign(\hat{g}, x)$ scores how well a group aligns with expert-defined criteria, and features in no group score $0$.

This formulation has two key properties: it is \emph{duplication-invariant} at optimality (repeating a feature group does not change the per-feature score) and it \emph{encourages diversity} (adding a new feature group that does not overlap with existing groups is always beneficial to the overall score, incentivizing broader coverage of the input space).

\begin{figure*}[t]
    \centering
    \includegraphics[width=\textwidth]{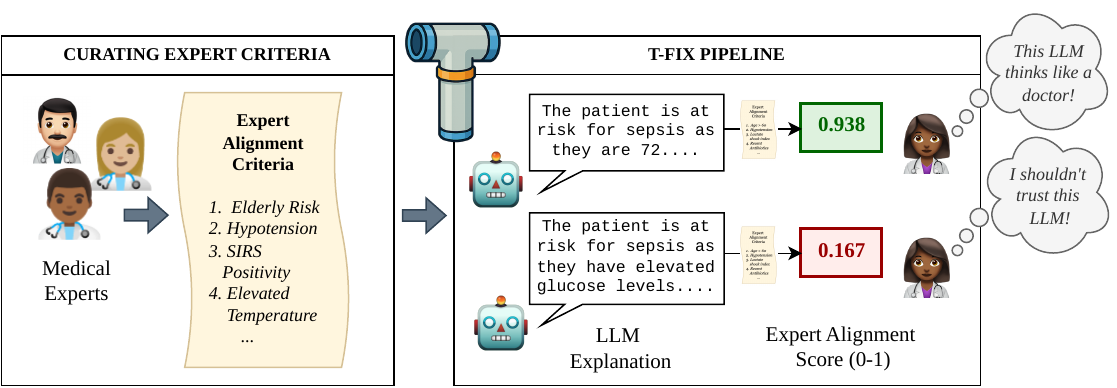}
    \caption{An overview of the \dataset construction process. For each dataset, we first work with domain experts to establish expert-alignment criteria --- features deemed important by domain experts for a specific task. These criteria form the basis of the \dataset evaluation pipeline, which processes an LLM-generated explanation to output an expert-alignment score. A high score suggests the explanation reflects reasoning aligned with domain experts, while a low score indicates the explanation may rely on aspects that experts would deem irrelevant.}
    \label{fig:criteria}
\end{figure*}

\subsection{Expert Criteria for Evaluating Explanations}
\label{sec:expert_criteria}

A straightforward way to generalize $\fixscore$ to natural language is to define $\featuregroupalign$ with respect to expert criteria also in natural language. Following \citet{jin2024fix}, we represent domain knowledge as a set of expert-defined criteria against which free-form explanations are scored, moving beyond example-level rubrics used in prior work~\cite{yang2026healthscorescalablerubricsimproving, li2025counselbenchlargescaleexpertevaluation} to reusable, task-level criteria that apply across all explanations for a given task.

\textbf{What a criterion looks like.} Each criterion $k \in K$ encodes a single piece of domain knowledge that experts use when reasoning about a task. For sepsis classification, one such criterion is:
\begin{quote}
\vspace{-0.25cm}
\texttt{\small Advanced age (over 65 years) increases susceptibility to rapid sepsis progression and higher mortality after infection.}
\vspace{-0.2cm}
\end{quote}
A complete criteria set $K$ captures the concepts, evidence, and inference patterns experts consider relevant for a prediction --- e.g., the features a clinician weighs when asking ``Will this patient develop sepsis in the next 12 hours?'' or the spectral signatures an astrophysicist looks for when asking ``What kind of supernova produced these wavelengths?'' An explanation is more expert-aligned the more its claims correspond to the criteria in this set.

\textbf{Developing the criteria.} For each of our seven domains (see Figure~\ref{fig:datasets}), we work with relevant domain experts --- five medical experts (surgeons and clinicians), three cosmologists (astrophysicists and astronomers), and two psychologists --- to assemble a criteria set. To make efficient use of their time, we seed each set using o3 deep research~\citep{openai2025o3o4minicard} to perform a literature review. Each prompt includes a task description, examples, and instructions to generate criteria, accompanied by citations from academic literature. We then present this seeded list to the experts and ask them to (1) remove incorrect or irrelevant criteria, (2) add important criteria that were missed, and (3) iteratively refine the list until it captures the field's knowledge. Experts left approximately 90\% of the seeded criteria unchanged. All prompt templates are in our GitHub repository, with final criteria in Appendix~\ref{sec:dataset-details}.

\textbf{No forced consensus between experts.} Within a domain, experts may disagree about which criteria matter most. Though our experts reached consensus, our pipeline is designed so criteria are modular and swappable --- alignment scores can be tailored to the priorities of the specific expert(s) evaluating an explanation, rather than forcing consensus. This follows the broader view that explanation needs vary across users, and that one-size-fits-all explanations are insufficient~\citep{sokol2020one,arya2019one}.

\subsection{T-FIX: Automatic and Generalizable Expert Alignment for Text}

However, it is not straightforward to apply the remainder of the FIX formulation to free-form text. Feature groups are, by construction, grounded, deduplicated, and pre-grouped; textual explanations satisfy none of these properties, so naively treating each sentence as a ``feature'' breaks the FIX formulation. A naive application of FIX to unstructured language would face the following problems:

\begin{itemize}
\vspace{-0.2cm}
\itemsep 0em
    \item \textbf{P1. Duplication invariance:} In FIX, repeating a feature group does not change the per-feature alignment score. Free-form text, however, can restate the same evidence in different words, and it is unclear which statements should be merged---without deduplication, LLMs would be rewarded for redundancy.
    \item \textbf{P2. Robustness to ungrounded content:} Feature groups in FIX are, by construction, binary masks over the input---every feature referenced is grounded. Free-form text, however, can contain claims that are speculative or ungrounded in the input (e.g., hallucinated facts). Such claims should receive no credit, even if they happen to align with expert reasoning.
    \item \textbf{P3. Group-level evaluation:} Alignment with expert reasoning often rests on compound evidence rather than any single statement, but free-form explanations lack inherent group structure. Meaningful claim groupings would be needed to penalize unaligned claims while granting full credit to those that contribute partially to a larger alignment.
\end{itemize}

The next section describes our automatic and generalizable pipeline for solving P1--P3.

%% file: sections/pipeline_new.tex
\section{The T-FIX Pipeline}

\begin{figure*}[t]
    \centering
    \includegraphics[width=\textwidth]{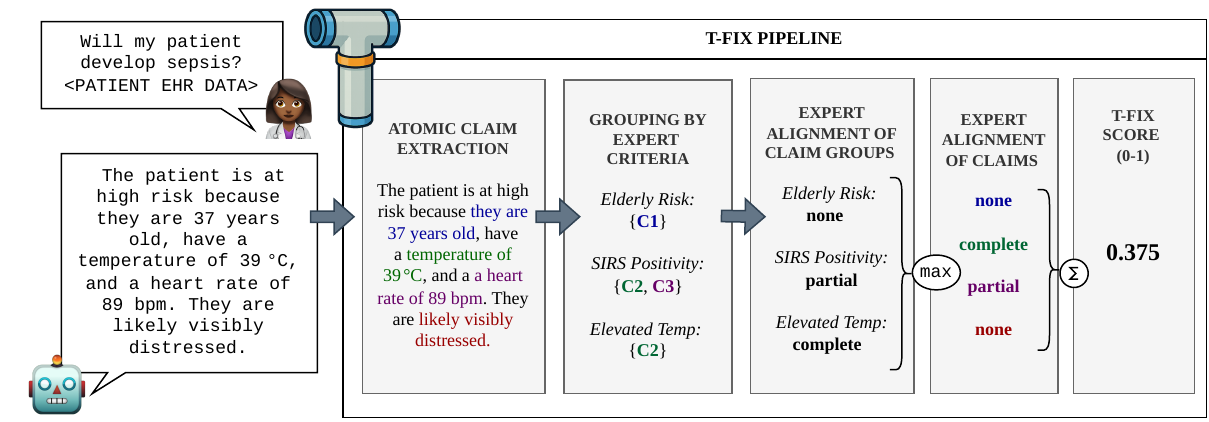}
    \caption{\textbf{Our \dataset pipeline for evaluating an LLM generated explanation.} We decompose the explanation into atomic claims, filter out ungrounded claims, and group the rest by expert criterion. Each group is scored as \textit{complete}, \textit{partial}, or \textit{none}. Each claim inherits the highest score among its groups (filtered or ungrouped claims receive \textit{none}), and the final score averages across all claims.}
    \label{fig:pipeline}
\end{figure*}

To solve P1--P3, we structure T-FIX as a four-stage pipeline (Figure~\ref{fig:pipeline}). Stage~1 enforces duplication invariance with claim decomposition (P1), Stage~2 filters out ungrounded content (P2), and Stages~3--4 together deliver group-level evaluation (P3): Stage~3 constructs the missing group structure over claims, and Stage~4 specifies how group scores are aggregated back to the claim level.

\textbf{General \dataset Formulation.} Let $x$ denote the task input, $y$ the model's prediction, $e$ the model-generated explanation for $y$, and $K = \{k_1, \ldots, k_m\}$ an expert-defined criteria set. 
For each claim $c_i$ extracted from $e$, we measure how well it aligns with respect to $K$, $e$, $x$, and $y$ as follows:
\begin{equation}
\claimalign(i, e, x, y) = 
\begin{cases}
0, & \text{if } K[i] = \varnothing \\[6pt]
\displaystyle\max_{k \in K[i]} \big(\groupalign(e, x, y, k)\big), & \text{otherwise}
\end{cases}
\label{eq:claimalign}
\end{equation}
where $K[i] = \{k \in K : c_i \in G_k\}$ are the criteria whose groups contain claim $i$. Claims that are ungrounded (filtered in Stage~2) or not related to any criterion receive a score of $0$.
We then define
\begin{equation}
\score(e, x, y) = \frac{1}{n}\sum_{i=1}^{n} \claimalign(i, e, x, y).
\label{eq:score}
\end{equation}
Intuitively, $\score(e, x, y)$ measures the fraction of an explanation's reasoning aligned with expert judgment. Every stage uses an evaluator LLM (such as GPT-5-mini~\citep{singh2025openaigpt5card}), with prompts developed iteratively with domain experts. Human validation is described in Section~\ref{sec:validation}.



\subsection{Stage 1: Atomic Claim Extraction}
\label{sec:stage1}
This stage addresses \textbf{P1 (duplication invariance)}. The same evidence may appear in different phrasings---without deduplication, restating evidence multiple times produces redundant claims that each receive credit, inflating the score (Equation~\ref{eq:score}).
To deduplicate repeated evidence, we decompose each explanation $e$ into \emph{atomic claims}: self-contained statements each conveying a single, context-independent fact. Semantically equivalent restatements are merged, so each piece of evidence is counted once. 
The evaluator LLM extracts these claims following standard claim decomposition techniques~\cite{wanner2024closer, gunjal2024molecularfactsdesideratadecontextualization}, yielding $C = \{c_1, \ldots, c_n\} = \mathsf{decompose}(e)$.

\subsection{Stage 2: Ungrounded Claim Filtering}
\label{sec:stage2}
This stage addresses \textbf{P2 (robustness to ungrounded content)}. Unlike feature groups in FIX, which reference input features by construction, free-form text can contain ungrounded claims (e.g., ``the patient appears visibly distressed'' when no such observation exists in the data).
The evaluator LLM retains only claims (1) grounded in $x$ and (2) directly explain \textit{why} the model predicted $y$, yielding $\widetilde{C} = \{c \in C : \mathsf{relevant}(c, x, y)\}$ (on average 76.8\% of claims pass).
Ungrounded claims receive $\claimalign(i, e, x, y) = 0$ (Equation~\ref{eq:claimalign}), thus strictly penalizing $\score$ when present.

\subsection{Stage 3: Claim Grouping and Scoring}
\label{sec:stage3}
\label{sec:claim_grouping}
This stage addresses \textbf{P3 (group-level evaluation)}. Unlike FIX, where explanatory units are pre-defined feature groups, free-form text has no natural grouping.
Evaluating all claims jointly against each criterion fails when claims supporting different criteria act as irrelevant noise for one another, penalizing diverse reasoning; evaluating claims individually fails for compositional criteria requiring multiple supporting facts.
We therefore construct \emph{criterion-level} groups: since criteria are the atomic units of expert knowledge, they provide a natural grouping axis. For each criterion, the evaluator LLM selects all related claims, so that only relevant claims are scored against it:
$G_k \coloneqq \mathsf{related}(\widetilde{C}, k) \subseteq \widetilde{C}.$
A claim may belong to multiple groups, and criteria with no related claims receive empty groups.


\paragraph{Group Scoring.}
The evaluator LLM labels each group as \emph{complete}, \emph{partial}, or \emph{none} alignment (Table~\ref{tab:score-interpretation}), mapped to $\groupalign(e, x, y, k) \in \{1, 0.5, 0\}$. A 3-point scale based on expert feedback avoids unneeded granularity; treating Likert ratings as equal-interval is conventional~\citep{norman2010likert}.

\subsection{Stage 4: Claim-Level Aggregation}
\label{sec:stage4}
\label{sec:max_agg}
This stage combines criterion-level scores into $\claimalign$ and $\score$ (Equations~\ref{eq:claimalign}--\ref{eq:score}).
The use of $\max$ contrasts with FIX's averaging, which is appropriate when feature groups are the explanatory units being evaluated. In our setting, criterion-level groups are an evaluation device, and Stage~3's soft grouping may assign a claim to criteria with varying relevance---e.g., elevated heart rate is strongly relevant to SIRS positivity but only tangentially to shock index. The aggregation should not conflate a limitation of the grouping with a deficiency of the claim.
We then formalize four desirable properties: (A1)~a claim need only align with one criterion to receive credit; (A2)~weaker tangential matches should not drag down that credit; (A3)~a claim matching only one criterion keeps that criterion's score; (A4)~no combination of partial matches should exceed the best match.
\begin{proposition}[Axiomatic $\claimalign$ Aggregation (informal)]
\label{prop:avgmax_informal}
    Let $f$ map any nonempty finite subset of $[0,1]$ to $[0,1]$, and define
$\claimalign_f(i, e, x, y) = f(S_i)$
where $S_i = \bigl\{\groupalign(e, x, y, k) : k \in K,\; c_i \in G_k\bigr\}$.
Then $f = \max$ uniquely satisfies A1--A4.
\end{proposition}

Proof is in Proposition~\ref{prop:avgmax} in Appendix~\ref{sec:proofs}. 

\begin{table}[t]
    \caption{Interpretation of alignment labels in scoring atomic claims against expert criteria.}
    \centering
    \small
    \renewcommand{\arraystretch}{1.3}
    \begin{tabularx}{\linewidth}{>{\centering\arraybackslash}m{1.5cm} X}
    \toprule
    \textbf{Label}   &  \textbf{Meaning}\\
    \midrule
    complete   &  The claims are all specific, directly relevant, and fully capture the meaning and intent of the expert criterion.\\ 
    partial   &  The claims partially support the expert criterion but lack key details, use vague language, are overly general, or contain noise.\\
    none   &  The claims reference something unrelated to the expert criterion, misinterprets the criterion's meaning, or contradict what is stated in the criterion.\\
    \bottomrule
    \end{tabularx}
    \label{tab:score-interpretation}
\end{table}

%% file: sections/pipeline_validation.tex
\section{Validation of Pipeline and T-FIXScore}
\label{sec:validation}

Given our pipeline relies on multiple curated prompts, we want to ensure that the extracted and filtered claims are accurate, and that the final alignment scores match domain expert intuition. To do this, we conduct a \textit{quantitative evaluation} via an annotation study, and a \textit{qualitative evaluation} via domain expert interviews.

\textbf{Expert-in-the-loop prompt development.} We met with our domain experts throughout the prompt development process for each step of our pipeline. We asked what they would do differently from the LLM, incorporated their feedback, and iterated on all prompts multiple times, until the experts were satisfied with the results. Our prompt iteration process is detailed in Appendix~\ref{app:prompt-engineering}. The final prompts for all three stages can be found in Appendix~\ref{sec:pipeline-prompts} and in our Github repository.

\textbf{Validating pipeline: annotation study.} To quantitatively validate the outputs at each stage, we manually annotate 35 examples (5 per domain), covering 295 extracted claims and 211 aligned claims. Six annotators participate, with two annotators per example.\footnote{Annotators are PhD students in machine learning trained extensively by relevant domain experts to evaluate LLM outputs.} See Appendix~\ref{app:annotation_study} for details. \cref{tab:validation} reports average accuracy and Cohen’s $\kappa$ across all seven \dataset datasets, indicating moderate-to-substantial agreement and supporting the reliability of our pipeline. Domain-level metrics appear in \cref{tab:domain_metrics}.

\textbf{Validating alignment scores: expert interviews.} To qualitatively validate the calculated $\claimalign$ scores and final $\score$ in stages 3 and 4 of our pipeline, we conduct validation studies with each group of domain experts for each task. The experts annotate and analyze 10-20 examples for their respective datasets, discussing noticed successes and failures. \textbf{\textit{Across domains, experts largely agreed with the LLM alignment labels, confirming that the evaluation framework reliably captures expert reasoning.}} Agreement was highest in domains with well-defined interpretive criteria (e.g., physics and psychology). Additional prompt iteration incorporating interview feedback was required to achieve high agreement in medicine, as multi-symptom, compositional medical statements were more complex. Overall, expert validation supports the interpretability and trustworthiness of the \dataset evaluation pipeline. We provide a summary of each domain expert's interview in Appendix~\ref{app:expert_validation}.

%% file: sections/experiments.tex
\section{Experiments on Benchmarking Frontier LLMs for Expert Alignment}
\label{sec:experiments}

\input{sections/datasets}

\textbf{Baselines.} We evaluate top frontier LLMs on \dataset using four prompting strategies: \textbf{\textit{(1) Vanilla}} (explain alongside the answer, no additional guidance), \textbf{\textit{(2) Chain-of-Thought}} (step-by-step reasoning before answering), \textbf{\textit{(3) Socratic Prompting}} (self-questioning to surface uncertainties), and \textbf{\textit{(4) Subquestion Decomposition}} (break the task into subquestions, answer each, then synthesize). Domain-specific prompts are in Appendix~\ref{sec:dataset-details}, with templates for the above prompting strategies in \cref{fig:baseline_prompts}. 
Results for GPT-5.2, GPT-5-mini, Gemini-2.5-Flash, Gemini-2.5-Pro, Claude-Opus-4.5, and Claude-Haiku-4.5 are in \cref{tab:baseline_scores}.~\footnote{We only select stable LLMs (i.e., not in preview mode) with vision support and context windows long enough to accommodate our time-series datasets. All models are accessed in January 2026.}

\textbf{Evaluator robustness.} Ablations across three evaluator LLMs (GPT-5-mini~\citep{singh2025openaigpt5card}, Gemini-2.5-Flash-Lite~\citep{comanici2025gemini25pushingfrontier}, Qwen2.5-VL-7B-Instruct~\citep{Qwen2.5-VL}) show that top-ranked and bottom-ranked models remain largely consistent across evaluators, despite minor variations in exact rankings and scores (Appendix~\ref{app:eval_ablation}).


%% file: sections/datasets.tex


\begin{figure*}[!t]
    \centering
    \includegraphics[width=\linewidth]{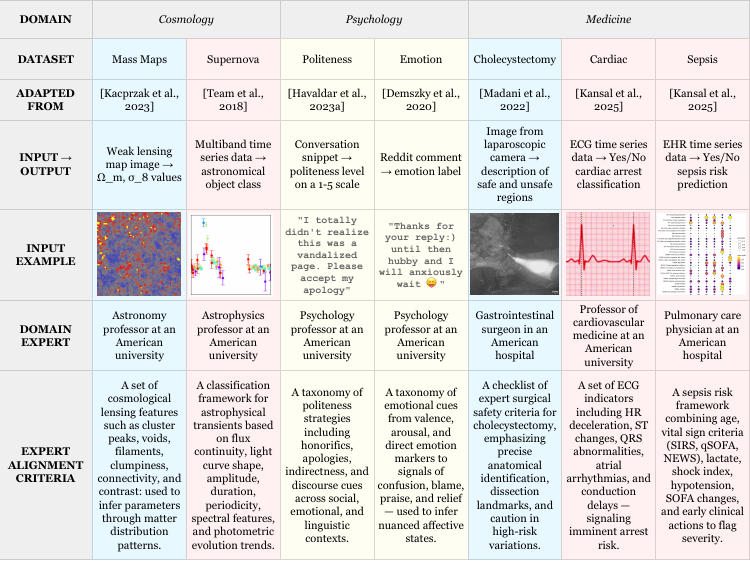}
    \caption{Overview of datasets and domains in \dataset. We evaluate LLM expert alignment across seven diverse scientific tasks, spanning cosmology, psychology, and medicine. The final row summarizes the expert alignment criteria used for scoring explanations in each domain. The column colors reflect dataset modality: blue = vision, yellow = language, and pink = time-series.}
    \label{fig:datasets}
\end{figure*}

\input{figures/baseline_scores}


\textbf{Datasets.}
\dataset aggregates seven open-source datasets, spanning the fields of cosmology, psychology, and medicine. To assess LLM explanations across multiple modalities, we include text, vision, and time-series datasets. We select these seven datasets due to the availability of domain experts willing to work with us for these tasks.
We select 100 examples from each dataset's test set (700 total), balanced across classes when possible, and cap at this size to manage API costs and keep \dataset accessible. An overview of all datasets is in Figure~\ref{fig:datasets}; per-dataset details are in Appendix~\ref{sec:dataset-details}.

%% file: figures/baseline_scores.tex
\begin{table*}[t]
\caption{Evaluating top LLMs on \dataset using GPT-5-mini. We report the average $\score$ across all examples in the dataset. Corresponding accuracies are in \cref{tab:accuracy_scores} and baseline prompting strategies are described in Section~\ref{sec:experiments}. We report $\pm$ standard deviation over 1000 bootstrap samples. In general, more sophisticated prompting strategies do not improve expert alignment of the explanations. }
\centering
\setlength{\tabcolsep}{2pt}
\scriptsize
\begin{tabular}{lccccccc}
\toprule
\multirow{2}{*}{} & \multicolumn{2}{c}{\textit{Cosmology}} & \multicolumn{2}{c}{\textit{Psychology}} & \multicolumn{3}{c}{\textit{Medicine}} \\
\textbf{Baseline} \hspace{0.2cm} & \textbf{Mass Maps}  & \textbf{Supernova} \hspace{0.1cm}    &  \hspace{0.1cm} \textbf{Politeness} & \textbf{Emotion} \hspace{0.1cm} & \hspace{0.1cm} \textbf{Cholecystectomy}     &  \textbf{Cardiac} & \textbf{Sepsis} \\
\midrule
\multicolumn{8}{l}{\textit{GPT-5.2-Pro \citep{openai_gpt52_2025}}} \\
\midrule
\textbf{Vanilla} & $0.901_{\pm 0.01}$ & $0.845_{\pm 0.01}$ & \best{$0.882_{\pm 0.01}$} & \best{$0.906_{\pm 0.01}$} & $0.538_{\pm 0.03}$ & $0.539_{\pm 0.02}$ & \best{$0.425_{\pm 0.03}$} \\
\textbf{Chain-of-Thought} & \bestoverall{$0.907_{\pm 0.01}$} & \best{$0.880_{\pm 0.01}$} & $0.876_{\pm 0.01}$ & $0.898_{\pm 0.01}$ & $0.632_{\pm 0.03}$ & \best{$0.587_{\pm 0.02}$} & $0.394_{\pm 0.03}$ \\
\textbf{Socratic Prompting} & $0.804_{\pm 0.01}$ & $0.714_{\pm 0.01}$ & $0.763_{\pm 0.02}$ & $0.726_{\pm 0.02}$ & \best{$0.656_{\pm 0.02}$} & $0.488_{\pm 0.02}$ & $0.381_{\pm 0.02}$ \\
\textbf{SubQ Decomposition} & $0.812_{\pm 0.01}$ & $0.727_{\pm 0.01}$ & $0.857_{\pm 0.01}$ & $0.774_{\pm 0.02}$ & $0.639_{\pm 0.02}$ & $0.489_{\pm 0.02}$ & $0.356_{\pm 0.02}$ \\
\midrule
\multicolumn{8}{l}{\textit{GPT-5-mini \citep{singh2025openaigpt5card}}} \\
\midrule
\textbf{Vanilla} & \best{$0.844_{\pm 0.01}$} & \bestoverall{$0.891_{\pm 0.01}$} & \best{$0.875_{\pm 0.02}$} & \best{$0.867_{\pm 0.02}$} & \best{$0.757_{\pm 0.02}$} & \best{$0.522_{\pm 0.02}$} & $0.410_{\pm 0.03}$ \\
\textbf{Chain-of-Thought} & $0.759_{\pm 0.01}$ & $0.876_{\pm 0.01}$ & $0.767_{\pm 0.02}$ & $0.830_{\pm 0.02}$ & $0.609_{\pm 0.03}$ & $0.507_{\pm 0.02}$ & $0.395_{\pm 0.03}$ \\
\textbf{Socratic Prompting} & $0.729_{\pm 0.01}$ & $0.781_{\pm 0.01}$ & $0.641_{\pm 0.02}$ & $0.690_{\pm 0.02}$ & $0.506_{\pm 0.03}$ & $0.444_{\pm 0.02}$ & $0.399_{\pm 0.02}$ \\
\textbf{SubQ Decomposition} & $0.720_{\pm 0.02}$ & $0.838_{\pm 0.01}$ & $0.809_{\pm 0.02}$ & $0.846_{\pm 0.02}$ & $0.528_{\pm 0.03}$ & $0.443_{\pm 0.01}$ & \best{$0.419_{\pm 0.03}$} \\
\midrule
\multicolumn{8}{l}{\textit{Claude-Opus-4.5 \citep{anthropic2025claudeopus45}}} \\
\midrule
\textbf{Vanilla} & $0.774_{\pm 0.02}$ & \best{$0.781_{\pm 0.02}$} & $0.827_{\pm 0.02}$ & \bestoverall{$0.916_{\pm 0.02}$} & $0.753_{\pm 0.01}$ & \bestoverall{$0.675_{\pm 0.02}$} & $0.376_{\pm 0.02}$ \\
\textbf{Chain-of-Thought} & $0.784_{\pm 0.02}$ & $0.752_{\pm 0.02}$ & \best{$0.854_{\pm 0.02}$} & $0.912_{\pm 0.01}$ & \bestoverall{$0.772_{\pm 0.01}$} & $0.627_{\pm 0.02}$ & \best{$0.396_{\pm 0.02}$} \\
\textbf{Socratic Prompting} & \best{$0.835_{\pm 0.01}$} & $0.683_{\pm 0.01}$ & $0.802_{\pm 0.02}$ & $0.789_{\pm 0.02}$ & $0.647_{\pm 0.01}$ & $0.653_{\pm 0.02}$ & $0.390_{\pm 0.02}$ \\
\textbf{SubQ Decomposition} & $0.765_{\pm 0.01}$ & $0.736_{\pm 0.01}$ & $0.800_{\pm 0.02}$ & $0.821_{\pm 0.02}$ & $0.546_{\pm 0.01}$ & $0.620_{\pm 0.02}$ & $0.386_{\pm 0.02}$ \\
\midrule
\multicolumn{8}{l}{\textit{Claude-Haiku-4.5 \citep{anthropic2025claudehaiku45}}} \\
\midrule
\textbf{Vanilla} & $0.742_{\pm 0.01}$ & $0.630_{\pm 0.02}$ & \best{$0.873_{\pm 0.02}$} & $0.881_{\pm 0.02}$ & $0.677_{\pm 0.01}$ & \best{$0.529_{\pm 0.02}$} & $0.355_{\pm 0.02}$ \\
\textbf{Chain-of-Thought} & \best{$0.781_{\pm 0.01}$} & $0.640_{\pm 0.02}$ & $0.871_{\pm 0.02}$ & \best{$0.887_{\pm 0.02}$} & \best{$0.722_{\pm 0.01}$} & $0.510_{\pm 0.02}$ & \best{$0.368_{\pm 0.02}$} \\
\textbf{Socratic Prompting} & $0.686_{\pm 0.01}$ & \best{$0.658_{\pm 0.02}$} & $0.838_{\pm 0.01}$ & $0.842_{\pm 0.01}$ & $0.687_{\pm 0.01}$ & $0.492_{\pm 0.02}$ & $0.300_{\pm 0.02}$ \\
\textbf{SubQ Decomposition} & $0.626_{\pm 0.02}$ & $0.537_{\pm 0.02}$ & $0.769_{\pm 0.02}$ & $0.841_{\pm 0.02}$ & $0.616_{\pm 0.01}$ & $0.505_{\pm 0.02}$ & $0.308_{\pm 0.02}$ \\
\midrule
\multicolumn{8}{l}{\textit{Gemini-2.5-Pro \citep{comanici2025gemini25pushingfrontier}}} \\
\midrule
\textbf{Vanilla} & $0.801_{\pm 0.01}$ & \best{$0.715_{\pm 0.02}$} & $0.843_{\pm 0.02}$ & \best{$0.888_{\pm 0.02}$} & \best{$0.697_{\pm 0.01}$} & \best{$0.315_{\pm 0.02}$} & $0.511_{\pm 0.02}$ \\
\textbf{Chain-of-Thought} & \best{$0.812_{\pm 0.01}$} & $0.697_{\pm 0.02}$ & \best{$0.877_{\pm 0.02}$} & $0.873_{\pm 0.02}$ & $0.691_{\pm 0.02}$ & $0.264_{\pm 0.02}$ & \best{$0.525_{\pm 0.02}$} \\
\textbf{Socratic Prompting} & $0.671_{\pm 0.01}$ & $0.616_{\pm 0.02}$ & $0.682_{\pm 0.02}$ & $0.655_{\pm 0.02}$ & $0.670_{\pm 0.01}$ & $0.281_{\pm 0.02}$ & $0.466_{\pm 0.02}$ \\
\textbf{SubQ Decomposition} & $0.731_{\pm 0.01}$ & $0.635_{\pm 0.02}$ & $0.759_{\pm 0.02}$ & $0.836_{\pm 0.02}$ & $0.654_{\pm 0.01}$ & $0.221_{\pm 0.01}$ & $0.522_{\pm 0.02}$ \\
\midrule
\multicolumn{8}{l}{\textit{Gemini-2.5-Flash \citep{comanici2025gemini25pushingfrontier}}} \\
\midrule
\textbf{Vanilla} & \best{$0.836_{\pm 0.01}$} & $0.668_{\pm 0.02}$ & $0.891_{\pm 0.02}$ & \best{$0.900_{\pm 0.02}$} & $0.685_{\pm 0.01}$ & \best{$0.337_{\pm 0.02}$} & \bestoverall{$0.562_{\pm 0.02}$} \\
\textbf{Chain-of-Thought} & $0.816_{\pm 0.01}$ & $0.682_{\pm 0.02}$ & \bestoverall{$0.898_{\pm 0.02}$} & $0.894_{\pm 0.02}$ & \best{$0.695_{\pm 0.01}$} & $0.269_{\pm 0.02}$ & $0.531_{\pm 0.02}$ \\
\textbf{Socratic Prompting} & $0.790_{\pm 0.01}$ & $0.683_{\pm 0.02}$ & $0.831_{\pm 0.02}$ & $0.812_{\pm 0.02}$ & $0.620_{\pm 0.01}$ & $0.315_{\pm 0.02}$ & $0.484_{\pm 0.02}$ \\
\textbf{SubQ Decomposition} & $0.800_{\pm 0.01}$ & \best{$0.688_{\pm 0.02}$} & $0.816_{\pm 0.02}$ & $0.841_{\pm 0.02}$ & $0.688_{\pm 0.01}$ & $0.300_{\pm 0.02}$ & $0.506_{\pm 0.02}$ \\
\bottomrule
\end{tabular}
\label{tab:baseline_scores}
\end{table*}


%% file: sections/discussion.tex
We use \dataset to answer three questions about the expert alignment of frontier LLMs:
(1)~How expert-aligned are explanations from current frontier LLMs?
(2)~Is expert alignment simply explained by task accuracy?
(3)~What reasoning patterns characterize more expert-aligned explanations?
Our results show that current LLMs still struggle to produce expert-aligned explanations, that higher accuracy does not reliably imply higher expert alignment, and that broader coverage of expert criteria is associated with stronger alignment scores.

\begin{figure*}[t]
\centering
\begin{minipage}[t]{0.48\linewidth}
    \centering
    \includegraphics[width=0.95\linewidth]{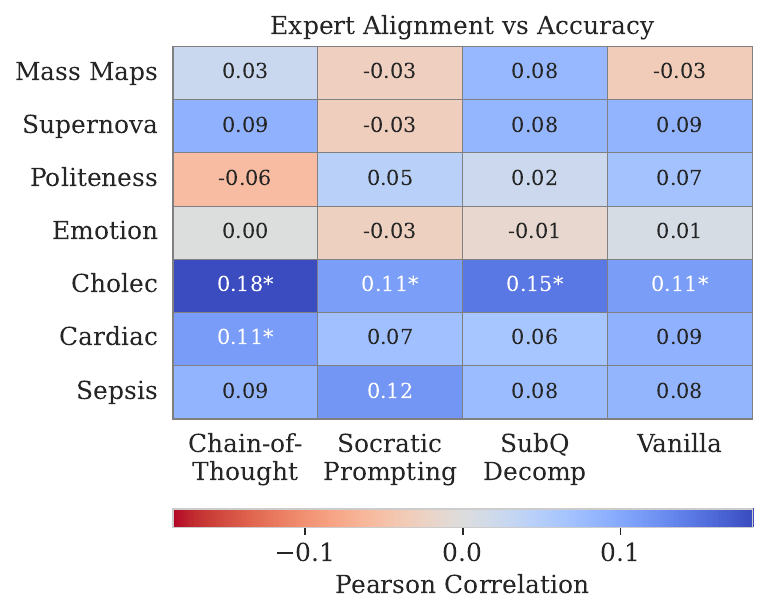}
    \caption{Expert Alignment vs. Accuracy Correlation Heatmap, averaged across all 6 models in \cref{tab:baseline_scores}. Blue indicates positive correlation, red is negative, gray is no correlation. $^*p < 0.05$.}
    \label{fig:correlation_heatmap}
\end{minipage}
\hfill
\begin{minipage}[t]{0.48\linewidth}
    \centering
    \includegraphics[width=\linewidth]{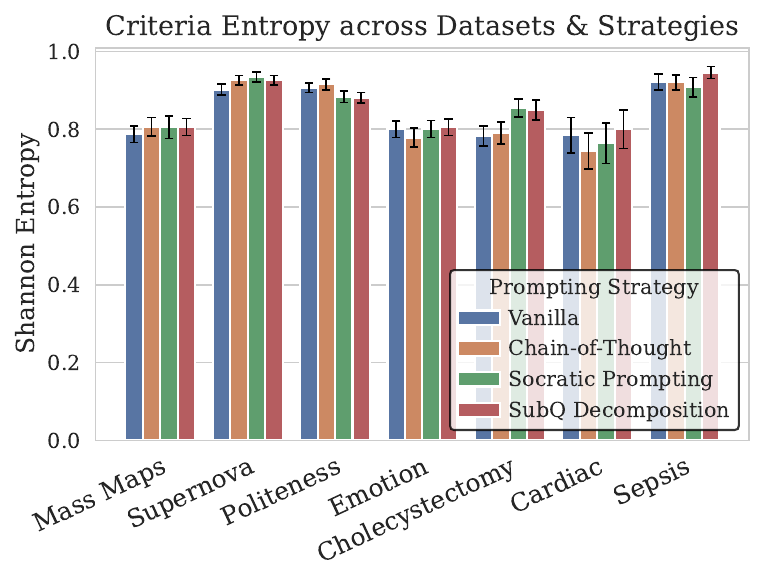}
    \caption{Shannon entropy (with bootstrap std, $n{=}1{,}000$, GPT-5-mini) of expert criteria  by prompting strategy. High entropy = diverse criteria; low entropy = repeating same few criteria.}
    \label{fig:entropy_histogram}
\end{minipage}

\end{figure*}

\subsection{Expert Alignment is an Open Problem for Frontier Models}
\label{subsec:benchmarking}
\cref{tab:baseline_scores} reports $\score$ across six frontier LLMs and four prompting strategies on all seven \dataset tasks. \textbf{\textit{Expert alignment varies sharply across domains}}: scores typically range from 0.63--0.91 in cosmology and psychology, but drop to 0.22--0.68 in medicine, with Sepsis rarely exceeding 0.55. This suggests that medical tasks, which require compositional, safety-critical clinical reasoning, pose a qualitatively harder challenge. \textbf{\textit{More sophisticated prompting strategies do not consistently help}}: Chain-of-Thought, Socratic Prompting, and Subquestion Decomposition sometimes improve and sometimes hurt alignment relative to Vanilla prompting, with no strategy dominating---the alignment gap cannot be closed by prompting alone. Finally, \textbf{\textit{no single model dominates across all domains}}, reinforcing that expert alignment is a broad challenge rather than a deficiency of any particular model.

\subsection{Expert Alignment is Orthogonal to Task Accuracy}
\label{subsec:accuracy}

\dataset evaluates explanation quality, but a natural question is whether expert alignment is simply explained by task accuracy---i.e., whether models that answer correctly also reason like experts. We compute Pearson correlations between $\score$ (\cref{tab:baseline_scores}) and accuracy (\cref{tab:accuracy_scores}) for each dataset and prompting strategy, averaged across all six models. \cref{fig:correlation_heatmap} shows that the vast majority of correlations are statistically insignificant or low in magnitude. A power analysis (Appendix~\ref{app:power_analysis}) confirms the benchmark is sufficiently powered to detect moderate correlations ($r \geq 0.30$), so the absence of significant results is not due to sample size. These results indicate that \textbf{\textit{we find no consistent association between prediction accuracy and expert alignment in current LLMs}}, suggesting expert alignment is a distinct, under-addressed dimension of LLM evaluation.

\subsection{Broader Expert-Criteria Coverage Correlates with Stronger Alignment}
\label{subsec:entropy}

Beyond measuring the proportion of expert-aligned claims, we ask whether LLMs that draw on a broader range of expert criteria also achieve higher alignment. Since individual explanations typically reference only 3--5 criteria, we analyze coverage at the dataset level: for each domain and prompting strategy, we compute the Shannon entropy of the distribution over criteria invoked across all examples. \cref{fig:entropy_histogram} shows that higher-$\score$ domains (e.g., Politeness, Emotion) exhibit more uniform coverage, while lower-performing domains (e.g., Cholecystectomy, Cardiac) show lower entropy, indicating models repeatedly invoke a narrow subset of criteria. This suggests \textbf{\textit{stronger expert-aligned explanations may require reasoning over a wider range of domain-specific criteria}}---though the relationship is correlational, leaving causation an open question.

%% file: sections/related.tex
\section{Related Work}
\label{sec:related}

\textbf{Evaluating LLM explanations at scale.} Common methods include self-generated explanations \cite{im2023evaluatingutilitymodelexplanations, zhao2023explainabilitylargelanguagemodels} or human-readable justifications \cite{camburu2018esnli}. To assess explanation quality and utility, faithfulness is analyzed \cite{jacovi2020faithfulness, zhou2021evaluating, agarwal2024faithfulness}. Human studies show mixed outcomes: explanations sometimes aid understanding \cite{hase2020evaluating, bansal2021whole}, but can also offer little value or cause over-trust \cite{wang2023fair}. A promising alternative is to use LLMs as automatic judges of explanation quality \cite{zheng2023judging, chen2024judge}, providing a scalable substitute for expensive human evaluation; we adopt this approach in \dataset.

\textbf{Domain \& expert alignment.}
Prior work incorporates domain knowledge through concept-based representations that constrain or structure predictions \cite{koh2020concept, kim2018tcav, chen2020concept, ghorbani2019ace, jin2024fix}, and through explanation supervision in NLP \cite{camburu2018esnli, bhatt2020explainable}. Existing efforts to evaluate expert alignment are intensive and small-scale \cite{kutbi2025evaluating, hyk2025queriescriteriaunderstandingastronomers}, while larger domain benchmarks rarely assess expert alignment beyond validation (e.g., Astro-QA \cite{li2025astronomical}, CounselBench \cite{li2025counselbenchlargescaleexpertevaluation}, Mentalchat16k \cite{xu2025mentalchat16k}) and rubric-based medical evaluations remain tied to fixed examples \cite{yang2026healthscorescalablerubricsimproving}. To our knowledge, no prior work automatically evaluates expert alignment and is generalizable to unseen explanations like \dataset does.

%% file: sections/conclusion.tex
\section{Conclusion}
\label{sec:conclusion}

We introduce \dataset, the first benchmark designed to evaluate LLM explanations for expert alignment across seven knowledge-intensive domains. Unlike existing evaluation frameworks, \dataset front-loads expert effort into a reusable framework that enables automatic expert-alignment scoring of unseen explanations. Our results show that generating explanations aligned with expert intuition remains an open challenge for current LLMs, orthogonal to their predictions accuracies.
Future work may include exploring instruction-tuning LLMs to generate explanations with strong expert alignment, extending \dataset to additional domains, and Human-Computer Interaction studies exploring how expert-aligned explanations affect real-world decision-making by practitioners.

\section*{Impact Statement}

This paper presents work whose goal is to advance the field of Machine
Learning. There are many potential societal consequences of our work, none
which we feel must be specifically highlighted here.

\section*{Limitations}

As with any LLM-based system, the quality of the outputs is dependent on the input prompt. \dataset is no exception -- though we spend a significant amount of time analyzing outputs and prompt iterating, we do a finite amount of prompt iteration. There is a chance our benchmark could be marginally improved with additional prompt iteration. We hope the issue of prompt dependency diminishes with future models that are more robust and less susceptible to tiny prompt ablations. 

While our evaluation pipeline currently uses GPT-5-mini for scoring, it is model-agnostic by design, and we encourage future work to apply or adapt the pipeline with other LLMs to improve robustness and reduce evaluator-model entanglement. For pipeline validation, we conduct a user study where we annotate 35 examples. Though the annotation results on this subset suggest our pipeline is accurate, this work could have benefited from a larger and more robust annotation study. In addition, we only have a small number of experts to validate the expert alignment criteria for each domain. Having additional experts would have been better to create a more robust expert criteria. We were constrained by the availability of domain experts.

Our experiments focus on a set of six models and four prompting strategies, and including additional models and strategies could provide a more comprehensive set of baseline results. Though many other high-performing LLMs and prompting techniques exist as of January 2026, we are conscious of budget and the environmental impact of running multiple experiments using \dataset. Additionally, LLMs are constantly changing, especially those that are company-owned and not open-source. This poses potential issues relating to the reproducibility of our baseline results as time progresses and advances are made.  

Finally, using LLMs in the domains we describe in \dataset, especially those relating to medicine, poses a unique set of risks and challenges. We do not advocate that LLMs should replace domain experts in these tasks; rather, \dataset should serve as a step towards experts being able to use LLMs in a reliable and trustworthy way. 




%% file: sections/acknowledgement.tex
\section*{Acknowledgment}
This research was partially supported by a gift from AWS AI to Penn Engineering's ASSET Center for Trustworthy AI, by ASSET Center Seed Grant, ARPA-H program on Safe and Explainable AI under the award D24AC00253-00, by NSF award CCF 2442421, and by funding from the Defense Advanced Research Projects Agency's (DARPA) SciFy program (Agreement No. HR00112520300). The views expressed are those of the author and do not reflect the official policy or position of the Department of Defense or the U.S. Government.

%% file: sections/appendix.tex
\appendix

\setcounter{table}{0}
\setcounter{figure}{0}
\renewcommand{\thetable}{A\arabic{table}}
\renewcommand{\thefigure}{A\arabic{figure}}

\input{sections/proofs}

\input{sections/power_analysis}

\section{Extending \dataset to a New Domain}
\label{sec:new-domain}
Though \dataset covers a wide range of knowledge-intensive settings, it can easily be extended to additional domains. 

A key contribution of the \dataset benchmark is the framework: we create a pipeline to score any free-form text explanation for expert alignment given a set of expert criteria. Additionally, we iterate extensively on all our prompt templates to ensure all \dataset users need to do is input their task-specific details and perform no additional prompt engineering for good results.   

To add a new domain to \dataset, we advise you to follow these steps:

\begin{enumerate}
\itemsep 0em
    \item \textbf{Generate criteria:} Use the deep research prompt template shown in \cref{fig:deep_research_prompts} to generate a list of expert alignment criteria for your domain. Optionally, have a domain expert vet the generated criteria.
    \item \textbf{Modify prompts:} Modify the prompt templates outlined in \cref{fig:stage1-prompt}, \cref{fig:stage2A-prompt}, \cref{fig:stage2B-prompt}, and \cref{fig:stage3-prompt} with your task description, few-shot examples, and generated expert criteria.
    \item \textbf{Run \dataset:} Plug in your prompts for each stage of the pipeline and run \dataset on your dataset! 
\end{enumerate}

We encourage you to contact the authors of this work if you need additional assistance setting up your custom domain.

\section{Annotation Study Setup}
\label{app:annotation_study}
\textbf{Annotator qualifications.} Annotators are PhD students in machine learning and the authors of this paper. Each author annotating for a scientific domain has had multiple meetings with that domain's experts, and been taught exactly how to evaluate an LLM's explanation for expert alignment. Note that this annotation study is meant to support the output of the \dataset pipeline \textit{in conjunction with} the qualitative study described in Appendix~\ref{app:expert_validation}.

Annotators evaluate each stage using a unified labeling scheme: (A) fully accurate, (B) partially accurate, or (C) incorrect. 

For atomic claim extraction, they review the explanation and its extracted claims, labeling outputs as (A) correct decomposition, (B) 1–2 claims missing/incorrect, or (C) 3+ claims missing/incorrect. 

For filtering, they assess whether claims were appropriately kept or removed, labeling results as (A) all correct, (B) 1–2 incorrect, or (C) 3+ incorrect. 

For expert alignment scoring, annotators review the grouped claims and expert criteria, labeling each as (A) consistent with expert-defined criteria, (B) borderline, or (C) inconsistent with expert-defined criteria.

These labels are then mapped to accuracy scores of 1.0, 0.5, and 0.0, respectively. \cref{tab:validation} reports average accuracy and Cohen’s $\kappa$ across all seven \dataset datasets, indicating moderate-to-substantial agreement and supporting the reliability of our pipeline. Domain-level metrics appear in \cref{tab:domain_metrics}.

\begin{table}[t]
    \caption{Pipeline validation: Average accuracy for $\mathcal{N}$ samples across all \dataset domains and annotator agreement -- Cohen's $\kappa$ for each stage in our pipeline. Domain-specific statistics are provided in \cref{tab:domain_metrics}.}
    \centering
    \small
    \renewcommand{\arraystretch}{1.3}
    \begin{tabular}{p{3cm}rrr}
    \toprule
    \textbf{Pipeline Stage} & \textbf{$\mathcal{N}$} & \textbf{Accuracy} & \textbf{Cohen's $\kappa$} \\
    \midrule
    Claim Extraction & 35  & 0.943 & 0.717 \\
    Filtering \& Grouping & 295 & 0.871 & 0.402 \\
    Expert Alignment & 211 & 0.923 & 0.405 \\
    \bottomrule
    \end{tabular}
    \label{tab:validation}
\end{table}

\input{figures/accuracy_scores}

\input{figures/domain_metrics}

\input{figures/ablation_scores}

\input{figures/claim_examples}

\section{Ablation for Evaluation Models}
\label{app:eval_ablation}
To assess the robustness of our evaluation pipeline, we conduct ablations using different models as evaluators: closed-sourced GPT-5-mini~\citep{singh2025openaigpt5card}, Gemini-2.5-Flash-Lite~\citep{comanici2025gemini25pushingfrontier}, and an open-source model Qwen2.5-VL-7B-Instruct~\citep{Qwen2.5-VL}. We evaluate vanilla prompts on three datasets, each representing a different domain: Mass Maps (cosmology), Emotion (psychology), and Cholecystectomy (medicine).
\Cref{tab:evaluator_ablation} summarizes the results:
\begin{itemize}
\itemsep 0em
\item \textbf{Mass Maps (cosmology):} All three evaluators unanimously rank the explanations from Claude-Opus-4.5 and Claude-Haiku-4.5 as the least aligned with expert intuition, while GPT and Gemini models align better.
\item \textbf{Cholecystectomy (medicine):} All three evaluators rank two of GPT-5.2-Pro, GPT-5-mini, and Claude-Opus-4.5 as the top two models. Claude-Haiku-4.5 and Gemini-2.5-Flash consistently rank in the bottom two, while remaining models cluster in the middle.
\item \textbf{Emotion (psychology):} All evaluators assign high scores across generation models, leading to minimal differences in rankings. Claude-Opus-4.5 is consistently ranked as one of the top two models.
\end{itemize}
Overall, these results indicate that our evaluation pipeline is \textit{robust to the choice of evaluation model}.

\input{sections/expert_validation}

\section{Prompts for \dataset Pipeline}
\label{sec:pipeline-prompts}
We show the prompts for Stage 1, 2, and 3 in \cref{fig:stage1-prompt}, \cref{fig:stage2A-prompt}, \cref{fig:stage2B-prompt}, and \cref{fig:stage3-prompt}, respectively. These prompts show a high-level template that was used by all domains. In practice, authors iterated multiple times on each domain's prompts, experimenting with the instruction wording and few-shot examples that yielded the best possible results. 

\input{sections/prompt_engineering}

\section{Compute Resources}
\label{app:compute}

All experiments use commercial LLM APIs for explanation generation and 
evaluation, except for Qwen2.5-VL-7B-Instruct, which runs on a single 
NVIDIA A100 GPU. No model training or fine-tuning is performed. Each 
model--dataset--prompting configuration takes approximately 1 hour of 
wall-clock time, yielding roughly 168 hours total for the main generation 
experiments (6 models $\times$ 4 strategies $\times$ 7 datasets) plus 
additional time for the evaluation pipeline and ablations. The total API 
cost was approximately \$1,500 USD.



\input{figures/prompts/prompt_templates}
\input{figures/prompts/baseline_prompts}

\section{\dataset Datasets: Additional Details}
\label{sec:dataset-details}

\Cref{fig:datasets} shows an overview of the datasets.

\paragraph{Dataset licenses.} All datasets used in \dataset are publicly 
available and properly licensed. In cosmology, 
CosmoGrid~\citep{cosmogrid1} is released under CC-BY 4.0, and 
PLAsTiCC~\citep{theplasticcteam2018photometric} is released on Zenodo 
under CC-BY 4.0. In psychology, 
GoEmotions~\citep{demszky2020goemotions} is released under Apache 2.0, 
and the multilingual politeness 
dataset~\citep{havaldar2025comparingstyleslanguagescrosscultural} is 
released under a CC license on HuggingFace. In medicine, the 
cholecystectomy data~\citep{madani2022artificial}, derived from 
Cholec80~\citep{twinanda2016endonet} and 
M2CAI16~\citep{stauder2016tum}, is released under a CC license on 
HuggingFace, and MC-MED~\citep{kansal2025mcmed}, used for both cardiac 
arrest and sepsis prediction, is available on PhysioNet under the 
PhysioNet Credentialed Health Data License (v1.5.0). Our use of all 
datasets complies with their respective terms.

\begin{figure*}[!t]
    \centering
    \includegraphics[width=\linewidth]{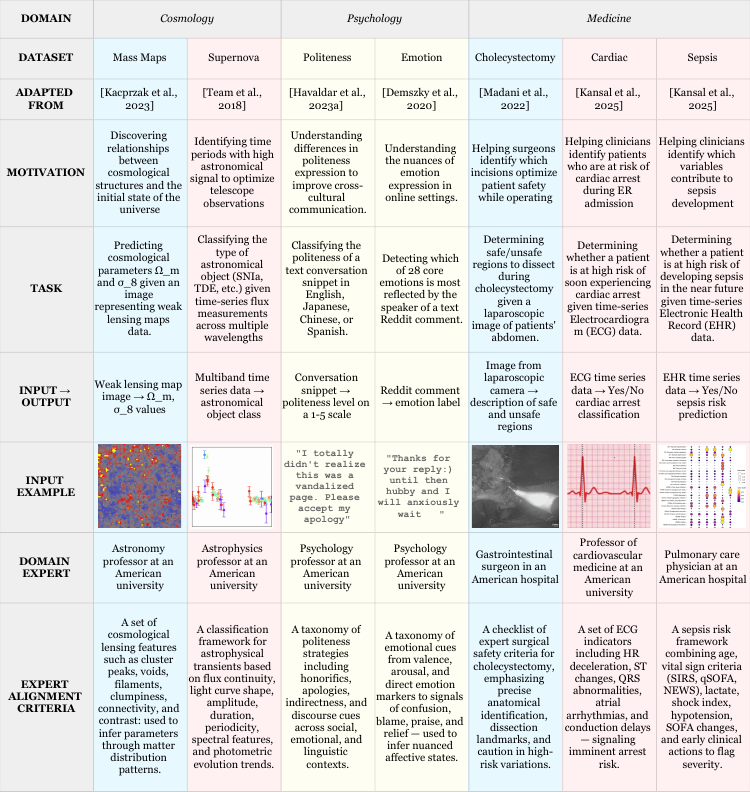}
    \caption{Overview of datasets and domains in \dataset. We evaluate LLM expert alignment across seven diverse scientific tasks, spanning cosmology, psychology, and medicine. For each dataset, we highlight the motivation, I/O format, representative example, and the key expert responsible for validation. The final row summarizes the expert alignment criteria used for scoring explanations in each domain. The column colors reflect dataset modality: blue indicates vision, yellow indicates language, and pink indicates time-series.}
    \label{fig:datasets-full}
\end{figure*}

\input{sections/datasets/mass_maps}

\input{sections/datasets/supernova}

\input{sections/datasets/politeness}

\input{sections/datasets/emotion}

\input{sections/datasets/cholec}

\input{sections/datasets/cardiac}

\input{sections/datasets/sepsis}

\input{sections/pipeline_details_app}

%% file: sections/proofs.tex

\section{Proofs}
\label{sec:proofs}
In this section we provide formal proof for the proposition stated in the main text.
\begin{restatable}[Axiomatic characterization of claim-level aggregation]{proposition}{propavgmax}
\label{prop:avgmax}
Let $f$ map any nonempty finite tuple of values in $[0,1]$ to a value in $[0,1]$, and define
$\claimalign_f(i, e, x, y) = f(S_i)$
where $S_i = \bigl(\groupalign(e, x, y, k)\bigr)_{k \in K[i]}$
is the tuple of group-level scores for all criteria containing claim $i$.
Then $f = \max$ is the unique function satisfying:
\begin{enumerate}[label=\textup{(A\arabic*)}]
    \item \textbf{Sufficient contribution.}
    If $\exists\, k \in K[i]$ such that $\groupalign(e, x, y, k) = 1$,
    then $f(S_i) = 1$.
    \item \textbf{Independence from weaker matches.}
    For any $s \in [0,\, f(S_i)]$,\;
    $f\!\bigl(S_i \oplus (s)\bigr) \;\geq\; f(S_i)$,
    where $\oplus$ denotes tuple concatenation.
    \item \textbf{Identity.}
    $f\bigl((a)\bigr) = a$ for all $a \in [0,1]$.
    \item \textbf{Boundedness.}
    $f(S_i) \leq \max(S_i)$.
\end{enumerate}
\end{restatable}

\begin{proof}
We first show that $\max$ satisfies all four properties, then show it is the unique such function.

\medskip\noindent\textbf{$\max$ satisfies (A1)--(A4).}
\begin{itemize}
    \item \emph{(A1).} If any entry of $S_i$ equals $1$, then $\max(S_i) = 1$.
    \item \emph{(A2).} Appending $s \leq \max(S_i)$ gives $\max(S_i \oplus (s)) = \max(S_i)$.
    \item \emph{(A3).} $\max\bigl((a)\bigr) = a$.
    \item \emph{(A4).} $\max(S_i) \leq \max(S_i)$ holds trivially.
\end{itemize}

\medskip\noindent\textbf{Uniqueness.}
Let $f$ satisfy (A1)--(A4) and let $S_i = (s_1, \dots, s_n)$ be any nonempty finite tuple of values in $[0,1]$.
Without loss of generality, reorder entries so that $s_1 = \max(S_i) \geq s_2 \geq \dots \geq s_n$
(permissible since all four axioms are permutation-invariant).

\emph{Lower bound.}
By (A3), $f\bigl((s_1)\bigr) = s_1$.
Since $s_j \leq s_1 = f\bigl((s_1)\bigr)$ for each $j \geq 2$, we can append $s_2, s_3, \dots, s_n$ one at a time;
at each step (A2) guarantees $f$ does not decrease.
Therefore $f(S_i) \geq s_1 = \max(S_i)$.

\emph{Upper bound.}
By (A4), $f(S_i) \leq \max(S_i) = s_1$.

Combining both bounds, $f(S_i) = \max(S_i)$.
\end{proof}

\begin{remark}
\label{rem:min_mean_violation}
As concrete illustrations: $\min$ and $\mathrm{mean}$ both violate (A1) and (A2).
For $S_i = (1, 0)$, both return a value less than $1$ despite $1$ appearing in $S_i$, violating (A1).
Starting from $S_i = (1)$ and appending $s = 0$, both decrease the aggregate, violating (A2).
\end{remark}

%% file: sections/power_analysis.tex
\section{Power Analysis for Expert Alignment vs.\ Accuracy}
\label{app:power_analysis}

In \cref{subsec:accuracy}, we report that Pearson correlations between $\score$ and task accuracy are largely insignificant across datasets and prompting strategies. Here we verify that this null result reflects a genuine absence of association rather than insufficient statistical power.

\paragraph{Setup.} Each of the seven \dataset tasks contains $n = 100$ examples. For each task and prompting strategy, we compute the Pearson correlation $r$ between per-example $\score$ and a binary or continuous accuracy measure. We test significance using the Fisher $z$-transform:
\begin{equation}
    z = \frac{1}{2} \ln\!\left(\frac{1+r}{1-r}\right), \quad \text{with standard error } \mathrm{SE} = \frac{1}{\sqrt{n-3}}.
\end{equation}
For $n = 100$, $\mathrm{SE} = 1/\sqrt{97} \approx 0.1015$.

\paragraph{Power calculation.} Under a two-sided test at significance level $\alpha = 0.05$ (critical value $z_{\alpha/2} = 1.96$), we compute the power to detect a true population correlation $\rho$ as:
\begin{equation}
    \text{Power}(\rho) = 1 - \Phi\!\left(z_{\alpha/2} - \frac{z(\rho)}{\mathrm{SE}}\right) + \Phi\!\left(-z_{\alpha/2} - \frac{z(\rho)}{\mathrm{SE}}\right),
\end{equation}
where $\Phi$ is the standard normal CDF and $z(\rho) = \frac{1}{2}\ln\!\left(\frac{1+\rho}{1-\rho}\right)$ is the Fisher-transformed population correlation. \cref{tab:power_analysis} reports power at several effect sizes:

\begin{table}[h]
\centering
\caption{Statistical power to detect a true Pearson correlation $\rho$ with $n=100$ and $\alpha=0.05$ (two-sided).}
\label{tab:power_analysis}
\begin{tabular}{lcccccc}
\toprule
True $\rho$ & 0.10 & 0.20 & 0.25 & 0.30 & 0.35 & 0.40 \\
\midrule
Power & 0.17 & 0.52 & 0.71 & 0.86 & 0.95 & 0.99 \\
\bottomrule
\end{tabular}
\end{table}

At $\rho = 0.30$---conventionally a medium effect size---the test achieves approximately 86\% power, comfortably above the standard 80\% threshold. For $\rho \geq 0.35$, power exceeds 95\%. The benchmark is therefore well-powered to detect any practically meaningful correlation between accuracy and expert alignment. The consistent absence of significant results across seven tasks and four prompting strategies (\cref{fig:correlation_heatmap}) provides strong evidence that expert alignment and task accuracy are largely independent dimensions in current LLMs.

%% file: figures/accuracy_scores.tex
\begin{table*}[t]

\caption{Evaluating top LLMs on \dataset. We report the average performance (accuracy/MSE/IoU) of the LLM across all examples in the dataset. We report accuracy for classification tasks, MSE for regression tasks ($^*$), and IoU for the Cholecystectomy task ($^\dagger$). Baseline implementations are described in Section~\ref{sec:experiments}. We report $\pm$ standard deviation over 1000 bootstrap samples.}
\centering
\setlength{\tabcolsep}{2pt}
\def\arraystretch{1.1}
\scriptsize
\begin{tabular}{lccccccc}
\toprule
\multirow{2}{*}{} & \multicolumn{2}{c}{\textit{Cosmology}} & \multicolumn{2}{c}{\textit{Psychology}} & \multicolumn{3}{c}{\textit{Medicine}} \\
    \textbf{Baseline} \hspace{0.2cm} & \textbf{Mass Maps}  & \textbf{Supernova} \hspace{0.1cm}    &  \hspace{0.1cm} \textbf{Politeness} & \textbf{Emotion} \hspace{0.1cm} & \hspace{0.1cm} \textbf{Cholecystectomy}    &  \textbf{Cardiac} & \textbf{Sepsis} \\
\midrule
\multicolumn{8}{l}{\textit{GPT-5.2-Pro \citep{openai_gpt52_2025}}} \\
\midrule
    \textbf{Vanilla} & $0.060^*_{\pm 0.01}$ & $0.236_{\pm 0.04}$ & $0.801^*_{\pm 0.11}$ & $0.320_{\pm 0.05}$ & $0.014^\dagger_{\pm 0.01}$ & $0.612_{\pm 0.04}$ & $0.627_{\pm 0.05}$ \\
    \textbf{Chain-of-Thought} & $0.058^*_{\pm 0.01}$ & $0.218_{\pm 0.04}$ & $0.868^*_{\pm 0.12}$ & $0.340_{\pm 0.05}$ & $0.162^\dagger_{\pm 0.02}$ & $0.623_{\pm 0.04}$ & $0.618_{\pm 0.05}$ \\
    \textbf{Socratic Prompting} & $0.062^*_{\pm 0.01}$ & $0.218_{\pm 0.04}$ & $0.841^*_{\pm 0.12}$ & $0.300_{\pm 0.05}$ & $0.098^\dagger_{\pm 0.02}$ & $0.623_{\pm 0.04}$ & $0.637_{\pm 0.05}$ \\
    \textbf{SubQ Decomposition} & $0.061^*_{\pm 0.01}$ & $0.227_{\pm 0.04}$ & $0.871^*_{\pm 0.12}$ & $0.340_{\pm 0.05}$ & $0.146^\dagger_{\pm 0.02}$ & $0.623_{\pm 0.04}$ & $0.627_{\pm 0.05}$ \\
\midrule
\multicolumn{8}{l}{\textit{GPT-5-mini \citep{singh2025openaigpt5card}}} \\
\midrule
    \textbf{Vanilla} & $0.048^*_{\pm 0.00}$ & $0.191_{\pm 0.04}$ & $1.004^*_{\pm 0.15}$ & $0.410_{\pm 0.05}$ & $0.149^\dagger_{\pm 0.01}$ & $0.585_{\pm 0.04}$ & $0.624_{\pm 0.05}$ \\
    \textbf{Chain-of-Thought} & $0.046^*_{\pm 0.00}$ & $0.164_{\pm 0.04}$ & $1.048^*_{\pm 0.14}$ & $0.350_{\pm 0.05}$ & $0.067^\dagger_{\pm 0.01}$ & $0.577_{\pm 0.04}$ & $0.676_{\pm 0.05}$ \\
    \textbf{Socratic Prompting} & $0.047^*_{\pm 0.00}$ & $0.200_{\pm 0.04}$ & $1.018^*_{\pm 0.15}$ & $0.310_{\pm 0.05}$ & $0.018^\dagger_{\pm 0.01}$ & $0.585_{\pm 0.04}$ & $0.627_{\pm 0.05}$ \\
    \textbf{SubQ Decomposition} & $0.050^*_{\pm 0.00}$ & $0.164_{\pm 0.03}$ & $1.024^*_{\pm 0.15}$ & $0.340_{\pm 0.05}$ & $0.011^\dagger_{\pm 0.00}$ & $0.577_{\pm 0.04}$ & $0.608_{\pm 0.05}$ \\
\midrule
\multicolumn{8}{l}{\textit{Claude-Opus-4.5 \citep{anthropic2025claudeopus45}}} \\
\midrule
    \textbf{Vanilla} & $0.035^*_{\pm 0.00}$ & $0.082_{\pm 0.03}$ & $0.701^*_{\pm 0.08}$ & $0.300_{\pm 0.05}$ & $0.273^\dagger_{\pm 0.01}$ & $0.454_{\pm 0.05}$ & $0.618_{\pm 0.05}$ \\
    \textbf{Chain-of-Thought} & $0.036^*_{\pm 0.00}$ & $0.118_{\pm 0.03}$ & $0.814^*_{\pm 0.12}$ & $0.300_{\pm 0.05}$ & $0.287^\dagger_{\pm 0.01}$ & $0.469_{\pm 0.04}$ & $0.634_{\pm 0.05}$ \\
    \textbf{Socratic Prompting} & $0.035^*_{\pm 0.00}$ & $0.109_{\pm 0.03}$ & $0.708^*_{\pm 0.08}$ & $0.310_{\pm 0.05}$ & $0.261^\dagger_{\pm 0.02}$ & $0.462_{\pm 0.04}$ & $0.620_{\pm 0.05}$ \\
    \textbf{SubQ Decomposition} & $0.035^*_{\pm 0.00}$ & $0.082_{\pm 0.03}$ & $0.774^*_{\pm 0.08}$ & $0.300_{\pm 0.05}$ & $0.277^\dagger_{\pm 0.01}$ & $0.469_{\pm 0.04}$ & $0.618_{\pm 0.05}$ \\
\midrule
\multicolumn{8}{l}{\textit{Claude-Haiku-4.5 \citep{anthropic2025claudehaiku45}}} \\
\midrule
    \textbf{Vanilla} & $0.040^*_{\pm 0.00}$ & $0.036_{\pm 0.02}$ & $0.681^*_{\pm 0.11}$ & $0.270_{\pm 0.04}$ & $0.290^\dagger_{\pm 0.01}$ & $0.677_{\pm 0.04}$ & $0.647_{\pm 0.05}$ \\
    \textbf{Chain-of-Thought} & $0.039^*_{\pm 0.00}$ & $0.036_{\pm 0.02}$ & $0.748^*_{\pm 0.12}$ & $0.260_{\pm 0.04}$ & $0.284^\dagger_{\pm 0.01}$ & $0.669_{\pm 0.04}$ & $0.608_{\pm 0.05}$ \\
    \textbf{Socratic Prompting} & $0.049^*_{\pm 0.00}$ & $0.091_{\pm 0.03}$ & $0.671^*_{\pm 0.10}$ & $0.270_{\pm 0.05}$ & $0.303^\dagger_{\pm 0.01}$ & $0.677_{\pm 0.04}$ & $0.324_{\pm 0.05}$ \\
    \textbf{SubQ Decomposition} & $0.043^*_{\pm 0.00}$ & $0.055_{\pm 0.02}$ & $0.728^*_{\pm 0.11}$ & $0.240_{\pm 0.04}$ & $0.296^\dagger_{\pm 0.01}$ & $0.677_{\pm 0.04}$ & $0.412_{\pm 0.05}$ \\
\midrule
\multicolumn{8}{l}{\textit{Gemini-2.5-Pro \citep{comanici2025gemini25pushingfrontier}}} \\
\midrule
    \textbf{Vanilla} & $0.097^*_{\pm 0.01}$ & $0.227_{\pm 0.04}$ & $0.901^*_{\pm 0.14}$ & $0.280_{\pm 0.04}$ & $0.216^\dagger_{\pm 0.01}$ & $0.408_{\pm 0.04}$ & $0.598_{\pm 0.05}$ \\
    \textbf{Chain-of-Thought} & $0.101^*_{\pm 0.01}$ & $0.209_{\pm 0.04}$ & $0.971^*_{\pm 0.15}$ & $0.270_{\pm 0.04}$ & $0.219^\dagger_{\pm 0.01}$ & $0.392_{\pm 0.04}$ & $0.588_{\pm 0.05}$ \\
    \textbf{Socratic Prompting} & $0.093^*_{\pm 0.01}$ & $0.173_{\pm 0.04}$ & $0.858^*_{\pm 0.15}$ & $0.290_{\pm 0.05}$ & $0.195^\dagger_{\pm 0.02}$ & $0.415_{\pm 0.04}$ & $0.569_{\pm 0.05}$ \\
    \textbf{SubQ Decomposition} & $0.087^*_{\pm 0.01}$ & $0.155_{\pm 0.04}$ & $0.815^*_{\pm 0.15}$ & $0.273_{\pm 0.04}$ & $0.151^\dagger_{\pm 0.01}$ & $0.408_{\pm 0.04}$ & $0.608_{\pm 0.05}$ \\
\midrule
\multicolumn{8}{l}{\textit{Gemini-2.5-Flash \citep{comanici2025gemini25pushingfrontier}}} \\
\midrule
    \textbf{Vanilla} & $0.081^*_{\pm 0.01}$ & $0.082_{\pm 0.03}$ & $1.048^*_{\pm 0.18}$ & $0.310_{\pm 0.05}$ & $0.201^\dagger_{\pm 0.01}$ & $0.385_{\pm 0.04}$ & $0.588_{\pm 0.05}$ \\
    \textbf{Chain-of-Thought} & $0.089^*_{\pm 0.01}$ & $0.145_{\pm 0.03}$ & $0.881^*_{\pm 0.15}$ & $0.340_{\pm 0.05}$ & $0.203^\dagger_{\pm 0.01}$ & $0.385_{\pm 0.04}$ & $0.578_{\pm 0.05}$ \\
    \textbf{Socratic Prompting} & $0.084^*_{\pm 0.01}$ & $0.091_{\pm 0.03}$ & $0.821^*_{\pm 0.14}$ & $0.310_{\pm 0.05}$ & $0.105^\dagger_{\pm 0.01}$ & $0.392_{\pm 0.04}$ & $0.608_{\pm 0.05}$ \\
    \textbf{SubQ Decomposition} & $0.085^*_{\pm 0.01}$ & $0.118_{\pm 0.03}$ & $1.058^*_{\pm 0.17}$ & $0.320_{\pm 0.05}$ & $0.190^\dagger_{\pm 0.01}$ & $0.405_{\pm 0.04}$ & $0.549_{\pm 0.05}$ \\
\bottomrule
\end{tabular}

\label{tab:accuracy_scores}
\end{table*}

%% file: figures/domain_metrics.tex
\begin{table*}[ht]
\caption{Pipeline validation by domain. We report the mean accuracy for each stage of the pipeline and annotator agreement -- Cohen's $\kappa$.}
\centering
\def\arraystretch{1.2}
\small
\begin{tabular}{lrrrrrr}
\toprule
\multirow{3}{*}{\textbf{Domain}} &
\multirow{3}{*}{\shortstack[r]{\textbf{$\mathcal{N}$} \\ \textbf{generated} \\ \textbf{claims}}} &
\multirow{3}{*}{\shortstack[r]{\textbf{$\mathcal{N}$} \\ \textbf{aligned} \\ \textbf{claims}}} &
\multirow{3}{*}{\shortstack[r]{\textbf{Claim} \\ \textbf{Decomposition} \\ \textbf{Accuracy}}} &
\multirow{3}{*}{\shortstack[r]{\textbf{Filtering \&} \\ \textbf{Grouping} \\ \textbf{Accuracy}}} &
\multirow{3}{*}{\shortstack[r]{\textbf{Expert} \\ \textbf{Alignment} \\ \textbf{Accuracy}}} &
\multirow{3}{*}{\shortstack[r]{\textbf{Cohen's $\kappa$}}} \\
\\
\\
\midrule
\textit{Cosmology} \\
\midrule
\textbf{Mass Maps} & 66 & 48 & 0.900 & 0.826 & 0.979 & 0.4059 \\
\textbf{Supernova} & 74 & 62 & 0.950 & 0.892 & 0.903 & 0.4946 \\
\midrule
\textit{Psychology} \\
\midrule
\textbf{Politeness} & 72 & 58 & 0.950 & 0.931 & 0.914 & 0.6604 \\
\textbf{Emotion} & 70 & 44 & 1.000 & 0.929 & 0.943 & 0.6233 \\
\midrule
\textit{Medicine} \\
\midrule
\textbf{Cholecystectomy} & 134 & 92 & 1.000 & 0.851 & 0.902 & 0.4396 \\
\textbf{Cardiac} & 66 & 52 & 0.900 & 0.841 & 0.962 & 0.4845 \\
\textbf{Sepsis} & 108 & 66 & 0.900 & 0.852 & 0.894 & 0.3500 \\
\bottomrule
\end{tabular}

\label{tab:domain_metrics}
\end{table*}

%% file: figures/ablation_scores.tex
\begin{table*}[t]
\caption{\textbf{Model ablation study for evaluation pipeline.} All experiments are run with vanilla prompting. Across domains, \textit{model rankings are largely consistent across evaluators, indicating the robustness of our evaluation pipeline.} Comparison of base models (explanation generator LLMs) across evaluator models (pipeline LLMs) for one dataset per domain. Best per evaluator is in \textbf{bold}; second-best is \underline{underlined}.}
\centering
\setlength{\tabcolsep}{4pt}
\small
\begin{tabular}{lrrr}
\toprule
\textbf{Evaluation Model} & \textbf{Gemini-2.5-Flash-Lite} & \textbf{GPT-5-Mini} & \textbf{Qwen2.5-VL-7B-Instruct} \\
\midrule
\multicolumn{4}{l}{\textit{MassMaps}} \\
\midrule
\textbf{GPT-5.2-Pro} & \secondbest{$0.749 \pm 0.011$} & \best{$0.901 \pm 0.008$} & \secondbest{$0.653 \pm 0.020$} \\
\textbf{GPT-5-mini} & $0.644 \pm 0.014$ & \secondbest{$0.844 \pm 0.011$} & $0.570 \pm 0.017$ \\
\textbf{Claude-Opus-4.5} & $0.558 \pm 0.018$ & $0.774 \pm 0.017$ & $0.468 \pm 0.016$ \\
\textbf{Claude-Haiku-4.5} & $0.452 \pm 0.018$ & $0.742 \pm 0.015$ & $0.485 \pm 0.017$ \\
\textbf{Gemini-2.5-Pro} & \best{$0.793 \pm 0.013$} & $0.801 \pm 0.012$ & $0.629 \pm 0.017$ \\
\textbf{Gemini-2.5-Flash} & $0.678 \pm 0.014$ & $0.836 \pm 0.011$ & \best{$0.691 \pm 0.017$} \\
\midrule
\multicolumn{4}{l}{\textit{Cholecystectomy}} \\
\midrule
\textbf{GPT-5.2-Pro} & \secondbest{$0.681 \pm 0.022$} & $0.538 \pm 0.030$ & \best{$0.686 \pm 0.022$} \\
\textbf{GPT-5-mini} & \best{$0.777 \pm 0.015$} & \best{$0.757 \pm 0.016$} & $0.643 \pm 0.019$ \\
\textbf{Claude-Opus-4.5} & $0.665 \pm 0.011$ & \secondbest{$0.753 \pm 0.011$} & \secondbest{$0.666 \pm 0.014$} \\
\textbf{Claude-Haiku-4.5} & $0.625 \pm 0.012$ & $0.677 \pm 0.010$ & $0.594 \pm 0.012$ \\
\textbf{Gemini-2.5-Pro} & $0.666 \pm 0.012$ & $0.697 \pm 0.012$ & $0.617 \pm 0.014$ \\
\textbf{Gemini-2.5-Flash} & $0.660 \pm 0.012$ & $0.685 \pm 0.012$ & $0.594 \pm 0.012$ \\
\midrule
\multicolumn{4}{l}{\textit{Emotion}} \\
\midrule
\textbf{GPT-5.2-Pro} & $0.729 \pm 0.019$ & \secondbest{$0.906 \pm 0.012$} & $0.728 \pm 0.028$ \\
\textbf{GPT-5-mini} & $0.730 \pm 0.020$ & $0.867 \pm 0.019$ & $0.751 \pm 0.025$ \\
\textbf{Claude-Opus-4.5} & \secondbest{$0.796 \pm 0.019$} & \best{$0.916 \pm 0.016$} & \secondbest{$0.752 \pm 0.027$} \\
\textbf{Claude-Haiku-4.5} & $0.751 \pm 0.024$ & $0.881 \pm 0.015$ & \best{$0.807 \pm 0.022$} \\
\textbf{Gemini-2.5-Pro} & \best{$0.806 \pm 0.023$} & $0.888 \pm 0.018$ & $0.723 \pm 0.029$ \\
\textbf{Gemini-2.5-Flash} & $0.772 \pm 0.026$ & $0.900 \pm 0.016$ & $0.743 \pm 0.028$ \\
\bottomrule
\end{tabular}
\label{tab:evaluator_ablation}
\end{table*}

%% file: figures/claim_examples.tex
\begin{table*}[!h]
\caption{Expert-aligned claims (good and bad) across all \dataset domains, with corresponding alignment scores and provided reasoning.}
\centering
\small
\def\arraystretch{1.25}
\begin{tabular}{p{1.5cm}p{4.4cm}p{3cm}p{5cm}}
\toprule
\textbf{Domain} & \textbf{Claim} & \textbf{Score (Category)} & \textbf{Reasoning} \\
\midrule
\textit{Cosmology} \\
\midrule
\multirow{5}{*}{\textbf{Mass Maps}} & The prominence of red and yellow suggests a universe with significant matter fluctuations. & Complete (\textit{Density Contrast Extremes}) & Aligns well with the Density Contrast Extremes category, describing pronounced contrasts between dense and void regions, signaling high sigma\_8. \\
& The mix of colors, with significant gray areas but noticeable reds and yellows, suggests a moderate Omega\_m. & Partial (\textit{Connectivity of the Cosmic Web}) & Discusses both underdense and overdense regions, but doesn't specifically discuss connectivity or the degree of fragmentation or interconnection of the network.\\
\multirow{5}{*}{\textbf{Supernova}} & A prominent peak followed by a gradual decline in flux is characteristic of a type Ia supernova light curve. & Complete (\textit{Rise–decline rates}) & Describes a classic feature of type Ia supernovae, perfectly aligning with expert criteria on rise-and-decline rates. \\
& The variability does not display a clear periodicity. & None (\textit{Periodic light curves}) & Contradicts key characteristics of periodic light curves; highlights absence of periodic behavior. \\
\midrule
\textit{Psychology} \\
\midrule
\multirow{5}{*}{\textbf{Politeness}} & The use of the phrase ``seems defective'' introduces uncertainty and avoids definitiveness. & Complete (\textit{hedging \& tentative language}) & The phrase utilizes tentative language and is a clear example of hedging to reduce the assertive strength of a statement.\\
& The utterance is a straightforward description of information from a biology textbook. & None (\textit{First-Person Subjectivity Markers}) & Does not align as it describes objective reporting without the personal tone central to first-person subjectivity. \\
\multirow{5}{*}{\textbf{Emotion}} & This choice of description is likely intended to evoke a reaction of fear or caution. & Complete (\textit{Threat/Worry Language}) & The claim centers around evoking fear or caution, which directly maps to this category. \\
& The text conveys an objective statement. & None (\textit{Valence}) & The claim highlights an absence of emotional content, which does not align with the Valence category or any other expert emotion categories. \\
\midrule
\textit{Medicine} \\
\midrule
\multirow{6}{*}{\textbf{Cholecys-}} & The fat and fibrous tissue overlying Calot's triangle has been fully excised, exposing only two tubular structures. & Complete (\textit{Complete Triangle Clearance}) & Precisely describes complete clearance of Calot's triangle, perfectly matching expert criteria. \\
\textbf{tectomy} & The cystic plate is not visible due to dense adhesions, making the gallbladder-liver plane indistinct. & None (\textit{Cystic Plate Visibility}) & Describes failure to visualize the cystic plate, opposite of the criterion, leading to low alignment. \\
\multirow{7}{*}{\textbf{Cardiac}} & The irregularity in the ECG could indicate a dangerous arrhythmia, such as ventricular tachycardia or fibrillation. & Complete (\textit{Ventricular Tachyarrhythmias}) & Directly references hallmark arrhythmias like ventricular tachycardia/fibrillation, key indicators in the category. \\
& A skin lesion of the scalp is a condition not directly related to cardiac function. & None (\textit{Critical Illness – Sepsis/Shock}) & Lacks explicit signs of sepsis/shock. \\
\multirow{5}{*}{\textbf{Sepsis}} & Fever and high heart rate are potential signs of sepsis. & Complete (\textit{SIRS Positivity}) & References two SIRS criteria; strong and direct alignment with early sepsis identification guidelines. \\
& The patient's lab results show an increased platelet count. & None (\textit{SOFA Score Increase}) & SOFA score focuses on low platelet counts; increased count contradicts the criterion. \\
\bottomrule
\end{tabular}

\label{tab:expert_alignment_examples}
\end{table*}

%% file: sections/expert_validation.tex
\section{Expert Validation}
\label{app:expert_validation}

\input{sections/expert/mass_maps}

\input{sections/expert/supernova}

\input{sections/expert/politeness}

\input{sections/expert/emotion}

\input{sections/expert/cholec}

\input{sections/expert/cardiac}

\input{sections/expert/sepsis}

%% file: sections/expert/mass_maps.tex
\subsection{Mass Maps}

Experts largely agreed with the model’s alignment on simple, descriptive claims that focused on observable visual patterns rather than interpretive reasoning. For instance, statements such as “The map displays a noticeable amount of blue and gray with significant red clustering” or mentions of “yellow spots” as dense regions were seen as well-aligned because they accurately reflected direct features in the image without overinterpreting them. The model was particularly consistent in correctly associating visible peaks and clusters with the relevant “Lensing Peak (Cluster) Abundance” category, showing strong performance on basic lensing-related observations.

However, disagreements arose for more interpretive claims where the model either misclassified the category or underestimated alignment. For example, experts judged “The significant red clustering indicates areas with some dense structures” as completely aligned, while the model labeled it partial. Similarly, when the model associated “yellow spots” with fine-scale clumpiness instead of large cluster convergence, experts noted categorical confusion. 

At a higher level, experts appreciated the model’s handling of concrete, color- or peak-based features but found its phrasing vague when dealing with broader cosmological interpretations. They preferred precise terminology about structure type and scale, emphasizing that generic descriptions such as “noticeable fluctuations” failed to convey sufficient scientific specificity.

%% file: sections/expert/supernova.tex
\subsection{Supernova}

Experts agreed with alignment scores for claims describing basic observable behaviors, such as “The light curve is followed by a gradual decline” or “The multi-wavelength observation shows a subsequent decline across all wavelengths,” which correctly corresponded to the “Monotonic Flux Trend” category. These statements were straightforward and descriptive, and the system’s partial alignment assessments matched expert expectations.

Disagreement arose for claims that overinterpreted observational patterns. For example, “The decline across all wavelengths corroborates the classification as a Type Ia supernova” was given full alignment by the model, but experts argued it should be partial, as such behavior is characteristic of transient events generally, not specific to Type Ia supernovae. 

Experts praised the model’s ability to identify key distinguishing features, like “The light curve shows a rapid rise to a peak,” which effectively differentiates supernovae from other celestial phenomena. However, they were less satisfied with claims that only ruled out other classes (e.g., RR Lyrae, AGN, Mira) without affirmatively justifying the target classification. They stressed that negative reasoning --- describing what an instance is not --- does not equate to correctly identifying what it is.

%% file: sections/expert/politeness.tex
\subsection{Politeness}

Experts generally agreed with the model’s alignment on explicit and lexical politeness cues. For example, both the model and experts assigned full alignment to claims like “The speaker expresses appreciation by saying ‘thank you,’” and “The utterance softens the request through the word ‘please.’” These cases show that the model reliably recognizes direct politeness indicators that have stable, context-independent meanings.

Disagreement arose for subtler forms of politeness that rely on social context or pragmatic interpretation. Experts noted that the model often overestimated alignment for hedges and modal constructions, such as I think, maybe, or could you, which did not always convey genuine politeness signal. For instance, the model rated “The use of ‘we’ conveys solidarity” as fully aligned, while experts marked it partial because the “we” referred to an institution rather than shared identity.

At a higher level, experts appreciated the model’s consistency and strong performance on overt politeness expressions but emphasized its limited pragmatic sensitivity. Overall, experts found the LLM trustworthy for identifying surface politeness cues but less dependable when interpreting indirect, contextual, or culturally grounded politeness strategies.

%% file: sections/expert/emotion.tex
\subsection{Emotion}

Experts generally agreed with the model’s alignment for surface-level emotional cues. Claims referencing Emotion Words, Emojis, Expressive Punctuation, Humor Markers were accurately scored by both the model and experts. The model also showed strong reliability in labeling claims related to Gratitude and Praise. These results suggest that the model can be trusted to judge alignment accurately when the emotion is lexically or visually explicit.

Disagreement emerged in more nuanced emotional expressions where context determined the emotional interpretation. Experts noted that the model often overestimated alignment for Valence and Arousal categories, particularly when phrasing was neutral or ambiguous. Claims involving subtle or mixed emotions, such as those involving Relief or Affection, were rarely labeled partial, as the model treated any positive cue as a strong emotional expression.

At a broader level, experts were happy with the model’s precision in labeling but noted limited sensitivity to gradation and context. The system performed best when emotions were unambiguous and directly expressed, but faltered when affect was implied, ironic, or intertwined with multiple sentiments. 

%% file: sections/expert/cholec.tex
\subsection{Laparoscopic Cholecystectomy}

Experts generally agreed with the model when it judged purely positional claims as not aligned with expert reasoning. They emphasized that referring only to a portion of the screen does not constitute surgically meaningful evidence for safety or risk. For example, the claim “The unsafe region occupies the lower to center-right portion of the image” was considered uninformative, as it describes only screen location rather than anatomically or surgically relevant cues. In such cases, experts agreed that alignment scores correctly reflected the lack of expert-valid reasoning.

Disagreement arose when the model assigned alignment to an incorrect anatomical category or relied on implausible visual inferences. For instance, experts judged the statement “The lower to center-right portion of the image shows darker, more vascular tissue” as less related to the “Peritoneal plane” category and more indicative of possible porta hepatis or vascular structures—although they emphasized that these structures cannot be directly identified from the image. Experts also noted that some unsafe-zone criteria, such as using the cystic lymph node as a landmark, are not visible in all cases and therefore difficult to apply consistently. In several instances, they pointed out that the model hallucinated tissues or structures that were not actually present in the image.

At a higher level, experts appreciated that the model correctly rejected claims based solely on screen position as misaligned with surgical intuition. However, they found the model less reliable when reasoning about anatomical structures whose visibility or relevance was uncertain. They stressed that many key anatomical landmarks cannot be directly observed in laparoscopic views and that some structures are far more informative than others for safety assessment. In particular, they highlighted the cystic lymph node and the estimation of the gallbladder infundibulum as useful landmarks, while cautioning against overinterpreting poorly visible or anatomically ambiguous regions.

%% file: sections/expert/cardiac.tex
\subsection{Cardiac}

Experts found most alignment scores accurate and consistent with their own judgments, highlighting the model’s general reliability in this domain. They appreciated the structural clarity and systematic nature of the pipeline, noting that alignment scores often captured correct relationships between claims and expert categories. Nevertheless, they pointed out that while scores were often correct, the model’s reasoning behind them was sometimes incomplete or slightly off, revealing a disconnect between the correctness of the label and the explanatory rationale.

At a broader level, experts valued the coherence of the approach and the alignment consistency across claims. However, they criticized redundancy and excessive complexity in some claims. For example, two statements about ECG rhythm regularity were nearly identical, suggesting overgeneration. They also noted that some claims were too dense and should be broken into smaller, more specific parts. Additional concerns included inaccuracies in certain category definitions, such as labeling 30 years old as “advanced age”, and insufficient background context for some explanations. 

Overall, they viewed the system as strong in structure but in need of refinement in content granularity and contextual accuracy.

%% file: sections/expert/sepsis.tex
\subsection{Sepsis}

Experts agreed with most alignment scores, particularly those labeled partial, since the claims captured correct but incomplete aspects of the clinical reasoning process. For instance, “One risk factor for sepsis is advanced age” was appropriately labeled as partial under “Elderly Susceptibility,” as it was factually true but lacked the specific clinical threshold (age $\geq$ 65). Similarly, claims about single vital sign abnormalities were correctly labeled as partial since sepsis criteria like SIRS require multiple indicators.

However, experts disagreed when the model evaluated isolated claims that should have been considered in combination. Statements like “A high triage temperature indicates fever” and “An elevated respiratory rate is another risk factor” were accurate but insufficient alone to establish “SIRS Positivity,” leading to misalignment between formal criteria and contextual reasoning. Additionally, experts highlighted that the LLM would generate explanations that involved risk factors that are not the same as diagnostic criteria for sepsis --- while risk factors increase the likelihood of sepsis, there is no causal relationship.

At a higher level, experts appreciated the clarity of claims tied to objective measures such as qSOFA scores and age but highlighted a persistent tension between rule-based alignment and real-world clinical reasoning. They emphasized that diagnostic logic in practice is more flexible than rigid scoring systems, exposing a conceptual gap that the model did not fully capture.

%% file: sections/prompt_engineering.tex
\section{Framework Development Process}
\label{app:prompt-engineering}
In this section, we detail how we engineered all the prompts we tried for each of the 4 prompts in our \dataset pipeline, what worked and what did not work, and how expert feedback was incorporated throughout the process.

\subsection{Atomic claim extraction}
The atomic claim extraction step decomposes each model-generated explanation into a set of atomic, independently interpretable claims for downstream relevance filtering and expert alignment. We designed structured prompts for each domain that enforce atomicity, standalone interpretability, and faithfulness to the original explanation.

\textbf{What did not work.}
Given we began with the refined atomic claim extraction prompts from \citet{wanner2024closer}, this step needed little prompt engineering to work as expected.

\textbf{What worked.}
We found that explicitly defining atomicity, standalone interpretation, and faithfulness within the prompt substantially improved the quality and consistency of extracted claims. Providing concrete in-context examples of correctly decomposed explanations for each domain further stabilized output structure and reduced stylistic variability across runs.
Additionally, instructing the model to output only a list of claims with no additional commentary simplified downstream processing and facilitated automated relevance filtering and alignment.

\subsection{Relevancy filtering}
In the relevancy filtering step, we determine whether each extracted claim should be retained as supporting evidence for the model’s prediction. We prompted the LLM to label a claim as relevant only if it (i) is supported by the provided input and (ii) directly contributes to explaining the predicted outcome.
This formulation filters out unsupported statements, generic background knowledge, and tautological claims that restate the prediction without adding evidence, yielding a cleaner set of claims for downstream grouping and expert alignment.

\textbf{What did not work.}
Prompts that framed relevance in vague terms (e.g., ``important,'' ``useful,'' or ``related'') led to inconsistent decisions and frequent retention of statements that were not meaningful for the explanation --- restating the prediction, describing the input, describing the task, etc. We required multiple rounds of prompt iterations and a clear definition of relevance to get this stage to work with high accuracy and reliability.

\textbf{What worked.}
We found that defining relevance with two explicit criteria, (i) the claim must be supported by the input and (ii) it must help explain the model’s prediction, yielded the most consistent filtering behavior. Requiring the model to generate a reasoning to support its decision improved performance and made failure modes easier to diagnose during expert review. Providing a set of diverse examples further helped the model decide which claims were speculative, unsupported by the input, or not meaningful for the explanation.

\subsection{Step 3: Claim grouping}
This step required the highest number of iterations, as it is where we most incorporated domain expert feedback. We have broken down our process to the main components of prompt generation for expert knowledge.

\textbf{Identifying Expert Categories.}
First, we used o3 deep research \cite{openai2025o3o4minicard} to seed our list of expert categories for each domain. We verified the model was pulling from reputable, existing literature, and manually ensured all categories made semantic sense and covered the areas that the experts expected. Following this, there were multiple (at least 2 per domain) iterations of review with the domain experts themselves for the determination of what constitutes the ``ideal'' expert categories for each domain. Working with the experts, we facilitated iterations where categories were modified, removed, and replaced, until experts reached consensus and the categories were finalized. 


\textbf{What did not work.}
Our initial setup of this pipeline did not involve the claim grouping step at all; we evaluated alignment of individual claims and aggregated the results. However, our domain experts pointed out that expert alignment is often determined by the joint content of multiple claims: claims can depend on one another, and some criteria are inherently compositional. For example, SIRS positivity in sepsis diagnosis requires meeting at least two symptoms from a predefined list, which may be expressed across several separate claims. Incorporating this feedback, we refactored our pipeline to incorporate claim grouping before expert alignment. This way, all claims related to a category could be judged together.

Even after introducing claim grouping, early prompts presented their own problems. Initial prompts returned groupings that were too selective; claims rarely belonged to multiple categories and many categories had zero related claims. Without explicit instructions to consider all claims before selecting a subset, GPT-5-mini sometimes defaulted to selecting only the most obviously related claims and ignored supporting, but less obvious evidence. The initial versions also frequently included irrelevant or weakly related statements in some categories, while omitting subtle but important claims phrased indirectly or lacking explicit category keywords. Without enforced exhaustive consideration of the full claim set, the model oscillated between being too restrictive and too permissive depending on the category.

\textbf{What worked.}
After finalizing the expert criteria, we developed prompts that allowed us to group all relevant, filtered claims into criteria-level groups. We allowed the claims to belong to multiple criteria, as certain facts or statements in our explanations were referenced by multiple criteria. What worked well was curating a diverse set of examples with varying downstream groupings: most claims in a category, around 1/2 of the claims in a category, and no claims in a category. GPT-5-mini erred on the side of fewer claims in a category, so we included more examples of categories with large claim groupings to account for this. Upon encouraging GPT-5-mini to be more generous and inclusive with the grouping, we were satisfied with the results.

\subsection{Expert alignment of claim groups}
This step also required a high amount of iteration, and domain expert feedback was heavily used in creating our final alignment prompts.

\textbf{What did not work.}
The first attempt of our claim to expert category alignment mapping was straightforward and simple. For each claim, we assigned it to the category that it most aligned with according to the LLM, and then calculated alignment. If the claim was completely not aligned with any of the expert categories, then it was omitted. We originally used a 0 to 1 scale with increments of 0.1 to score how aligned a claim was to a given expert category, where 1 is completely aligned and 0 is not aligned at all. This turned out to be too fine-grained a measure because it was unclear and ambiguous how to determine the difference between claims that got scores of 0.4 and 0.6, for example. Moreover, numeric scores encouraged spurious precision and made it difficult for domain experts to interpret or validate the assignments. We also found that even with explicit definitions of the scale, the LLM's scoring was unstable across runs and sensitive to minor prompt changes.

Separately, the initial versions of the alignment prompt frequently resulted in the inclusion of too many claims per category, including irrelevant or weakly related statements. The LLM also tended to omit subtle but important claims when they were phrased indirectly or lacked explicit category keywords. Without explicit instructions to consider all claims before selecting a subset, the model sometimes defaulted to selecting only the most obvious claims and ignored secondary evidence. This required prompt revisions that enforced exhaustive consideration of the claim set and stricter inclusion criteria.

\textbf{What worked.}
Based on feedback from our experts, we added the claim grouping step (see above). This group-based formulation better captured how expert reasoning often relies on multiple supporting statements rather than isolated claims. After experimentation, we also determined that using discrete alignment labels: \textit{none}, \textit{partial}, and \textit{complete} alignments for each claim group was the most effective approach, because it provided clear semantic interpretation of how well a claim reflected an expert category. This formulation reduced ambiguity and improved consistency compared to numeric scoring schemes.

Examples of claim groups with partial alignment were especially important in helping GPT-5-mini handle borderline cases where claims were related to a category but lacked sufficient specificity. Similar to previous steps, requiring the LLM to provide a short justification for each alignment decision helped debug errors and improved performance.

%% file: figures/prompts/prompt_templates.tex
\begin{figure*}[ht]
  \centering
  \begin{baselinepromptbox}
You will be given a paragraph that explains <task description>. Your task is to decompose this explanation into individual claims that are:

Atomic: Each claim should express only one clear idea or judgment.
Standalone: Each claim should be self-contained and understandable without needing to refer back to the paragraph.
Faithful: The claims must preserve the original meaning, nuance, and tone.

Format your output as a list of claims separated by new lines. Do not include any additional text or explanations.

Here is an example of how to format your output:
INPUT: [example]
OUTPUT: [example]

Now decompose the following paragraph into atomic, standalone claims:
INPUT:
  \end{baselinepromptbox}
\caption{Prompt Template for Stage 1: Atomic Claim Extraction}
\label{fig:stage1-prompt}
\end{figure*}

\begin{figure*}[ht]
  \centering
  \begin{baselinepromptbox}
You will be given <description of input, output, and claim>

A claim is relevant if and only if:
(1) It is supported by the content of the input (i.e., it does not hallucinate or speculate beyond what is said).
(2) It helps explain why <task description>.

Return your answer as:
Relevance: <Yes/No>
Reasoning: <A brief explanation of your judgment, pointing to specific support or lack thereof>

Here are some examples:
[Example 1]
[Example 2]
[Example 3]

Now, determine whether the following claim is relevant to the given <input example>:
Input:
Output:
Claim:
  \end{baselinepromptbox}
\caption{Prompt Template for Stage 2A: Relevancy Filtering}
\label{fig:stage2A-prompt}
\end{figure*}

\begin{figure*}[ht]
  \centering
  \begin{baselinepromptbox}
You will be given <description of claims and expert categories>

Your task is to behave like a <title of domain expert> and identify which atomic claims are related to the given expert category.
We define "related" as claims that are topically relevant to the expert category and/or can be used to support the expert category.

Task description:
Input:
Output:

-----
Expert categories:
[list of categories and their descriptions]
-----

Here are some examples:
[Example 1]
[Example 2]
[Example 3]

Now identify which atomic claims are related to the given expert category:
Category: {}
Claims: {}
  \end{baselinepromptbox}
\caption{Prompt Template for Stage 2B: Claim Grouping}
\label{fig:stage2B-prompt}
\end{figure*}

\begin{figure*}[ht]
  \centering
  \begin{baselinepromptbox}
You will be given <task description + expert categories description>

Your task is as follows:
Rate how strongly the set of claims align with the category. Choose from complete, partial, or none.

Alignment explanations:
Complete: The claim is specific, directly relevant, and fully captures the meaning and intent of the expert category.
Partial: The claim partially refers to the expert category but lacks key details, uses vague language, is overly general, or contains noise.
None: The claim references something unrelated to the expert category, or misinterprets the category's meaning

-----
Expert categories:
[list of categories and their descriptions]
-----

Return your answer as:
Reasoning: <A brief explanation of why you judged the alignment rating as you did.>
Category Alignment Rating: <rating>

Here are some examples:
[Example 1]
[Example 2]
[Example 3]

Now, determine the alignment rating for the following expert category and set of claims:
Category: {}
Claims: {}

  \end{baselinepromptbox}
\caption{Prompt Template for Stage 3: Alignment Scoring}
\label{fig:stage3-prompt}
\end{figure*}

%% file: figures/prompts/baseline_prompts.tex
\begin{figure*}[ht]
  \centering
  \begin{baselinepromptbox}
You are an expert in <domain name>. You have a deep understanding of this subject. 
Your task is to behave like an <domain expert> and identify which criteria are important to consider for the following task:

Task description:
Input:
Output:

Here are some examples:
[Example 1]
[Example 2]
[Example 3]

Study these examples and fully understand the task. Now, research the field of <domain name> in order to determine a list of criteria that an expert <domain expert> would utilize if they were performing the above task.

Your output should be a list of expert criteria, each 1 sentence long, and citations from reputable academic sources to support each criteria. Feel free to have as many expert criteria as you deem necessary. The criteria should be clear, succinct and non-overlapping with each other. [Include any domain-specific information about the expert criteria]
  \end{baselinepromptbox}
\caption{Deep Research Prompt Template.}
\label{fig:deep_research_prompts}
\end{figure*}

\begin{figure*}[ht]
  \centering
  \begin{baselinepromptbox}
VANILLA
In addition to the answer, please provide 3-5 sentences explaining why you gave the answer you did.

CHAIN-OF-THOUGHT
To come up with the correct answer, think step-by-step. You should walk through each step in your reasoning process and explain how you arrived at the answer. Describe your step-by-step reasoning in 3-5 sentences. This paragraph will serve as the explanation for your answer.

SOCRATIC
To come up with the correct answer, have a conversation with yourself. Pinpoint what you need to know, ask critical questions, and constantly challenge your understanding of the field. Describe this question-and-answer journey in 3-5 sentences. This paragraph will serve as the explanation for your answer.

SUBQUESTION DECOMPOSITION
To come up with the correct answer, determine all of the subquestions you must answer. Start with the easiest subquestion, answer it, and then use that subquestion and answer to tackle the next subquestion. Describe your subquestion decomposition and answers in 3-5 sentences. This paragraph will serve as the explanation for your answer.

  \end{baselinepromptbox}
\caption{Baseline Prompting Strategies.}
\label{fig:baseline_prompts}
\end{figure*}

%% file: sections/datasets/mass_maps.tex
\input{figures/prompts/fig_massmaps_explanation_prompt}

\subsection{Mass Maps}

\textbf{Task.}
The goal is to predict two cosmological parameters—\Om{} and \seight{}—from a weak lensing map (also known as mass maps)~\citep{Abbott_2022}. 
These parameters characterize the early state of the universe.
Weak lensing maps can be obtained through precise measurement of galaxies~\citep{y3-massmapping,y3-shapecatalog}, but it is not yet known how to characterize \Om{} and \seight{}.
There are machine learning models trained to predict \Om{} and \seight{}~\citep{ribli2019weak,matilla2020weaklensing,Fluri_2022}, as well as interpretable models that attempt to find relations between interpretable features voids and clusters and \Om{} and \seight{}~\citep{you2023sumofparts}.
 We use data from CosmoGrid~\citep{cosmogrid1}, where inputs are single-channel, noiseless weak lensing maps of size (66, 66), and outputs are two continuous values corresponding to \Om{} and \seight{}.

\textbf{Data Selection \& Preprocessing.}
We randomly sampled 100 examples from the MassMaps test set. To ensure compatibility with LLMs like GPT-5-mini, which operate on a 32×32 patch size, we upsampled each image by a factor of 11 to preserve spatial detail and avoid patch-level compression. Instead of raw pixel values, we applied a colormap based on expert-defined intensity thresholds used to identify key cosmological features such as voids and clusters. Pixel intensities were scaled by standard deviations to emphasize meaningful variation. We found that larger, visually enhanced inputs reduced refusal rates from LLMs and encouraged more consistent responses.

\textbf{Explanation Prompt.}
Figure~\ref{fig:massmaps_prompt} shows the prompt used to generate LLM explanations for predicting \Om{} and \seight{}. We replace \texttt{[BASELINE\_PROMPT]} with one of four prompting strategies shown in \cref{fig:baseline_prompts}. The prompt includes a description of how pixel values are mapped to colors, as well as the valid ranges for \Om{} and \seight{}. Without this range, models tend to default to common values (e.g., 0.3 for \Om{}, 0.8 for \seight{}), reducing response variability.

\textbf{Expert Criteria.} The expert-validated criteria for expert alignment calculation are listed below:
\begin{enumerate}
\small
\itemsep 0em
\small
\itemsep 0em
    \item \textbf{Lensing Peak (Cluster) Abundance:} High peak count $\rightarrow$ higher \seight{}; clumpy halos more common.
\item \textbf{Void Size and Frequency:} Large, frequent voids $\rightarrow$ lower \Om{}; less overall matter.
\item \textbf{Filament Thickness and Sharpness:} Thick, sharp filaments track higher \seight{}; thin indicates lower.
\item \textbf{Fine-Scale Clumpiness:} Fine graininess signifies high \seight{}; smooth map implies lower.
\item \textbf{Connectivity of the Cosmic Web:} Interconnected web suggests higher \Om{}; isolated clumps imply lower.
\item \textbf{Density Contrast Extremes:} Strong density contrast denotes high \seight{}; muted contrast lower.
\end{enumerate}

%% file: figures/prompts/fig_massmaps_explanation_prompt.tex
\begin{figure*}[ht]
  \centering
  \begin{promptbox}
You are an expert cosmologist.
You will be provided with a simulated noisless weak lensing map,

Your task is to analyze the weak lensing map given, identify relevant cosmological structures, and make predictions for Omega_m and sigma_8.
Each weak lensing map contains spatial distribution of matter density in a universe. The weak lensing map provided is simulated and noiseless.
Omega_m captures the average energy density of all matter in the universe (relative to the total energy density which includes radiation and dark energy).
sigma_8 describes the fluctuation of matter distribution. 

When you analyze the weak lensing map image, note that the number is below 0 if it shows up as between gray and blue, and 0 is gray, and between 0 and 2.9 is between gray and red, and above 2.9 is yellow. The numbers are in standard deviations of the mass map.

Omega_m's value can be between 0.1 ~ 0.5, and sigma_8's value can be between 0.4 ~ 1.4.
Note that the weak lensing map given is a simulated weak lensing map, which can have Omega_m and sigma_8 values of all kinds.

[BASELINE_PROMPT]

The provided image is the weak lensing mass map for you to predict the cosmological parameters for.
Your response should be 2 lines, formatted as follows (without extra information):
Explanation: <explanation and reasoning, as described above, 3-5 sentences>
Prediction: Omega_m: <prediction for Omega_m, between 0.1 ~ 0.5, based on this weak lensing map>, sigma_8: <prediction for sigma_8, between 0.4 ~ 1.4, based on this weak lensing map>
  \end{promptbox}
\caption{MassMaps Explanation Prompt}
\label{fig:massmaps_prompt}
\end{figure*}

%% file: sections/datasets/supernova.tex
\input{figures/prompts/fig_supernova_explanation_prompt}
\subsection{Supernova}

\textbf{Task.}
The objective is to classify astrophysical objects using time-series data comprising observation times (Modified Julian Dates), wavelengths (filters), flux values, and corresponding flux uncertainties. We use data from the PLAsTiCC challenge~\citep{theplasticcteam2018photometric}, where the model must predict one of 14 astrophysical classes.

\textbf{Data Selection \& Preprocessing.}
We sampled 100 examples across the Supernova train, validation, and test sets, aiming for 7–8 instances per class to mitigate class imbalance. For rare classes with only one test set instance, we included all available examples from the validation and test sets, supplementing with training samples to meet the target count. For LLM input, we converted each raw time series into a multivariate time-series plot: time is on the x-axis, flux on the y-axis, error bars denote flux uncertainty, and point colors indicate different wavelengths.

\textbf{Explanation Prompt.}
Figure~\ref{fig:supernova_prompt} shows the prompt used to generate explanations for classifying astronomical objects. We replace \texttt{[BASELINE\_PROMPT]} with one of four prompting strategies shown in \cref{fig:baseline_prompts}. The prompt includes a description of the input plot as a multivariate time series and provides the full list of possible class labels to guide the model's predictions.

\textbf{Expert Criteria.} The expert-validated criteria for expert alignment calculation are listed below:
\begin{enumerate}
\small
\itemsep 0em
\small
\itemsep 0em
    \item \textbf{Contiguous non-zero flux:} Contiguous non‑zero flux segments confirm genuine astrophysical activity and define the time windows from which transient features should be extracted.
    \item \textbf{Rise–decline rates:} Characteristic rise‑and‑decline rates—such as the fast‑rise/slow‑fade morphology of many supernovae—encode energy‑release physics and serve as strong class discriminators.
    \item \textbf{Photometric amplitude:} Peak‑to‑trough photometric amplitude separates high‑energy explosive events (multi‑magnitude outbursts) from low‑amplitude periodic or stochastic variables.
    \item \textbf{Event duration:} Total event duration, measured from first detection to return to baseline, distinguishes short‑lived kilonovae and superluminous SNe from longer plateau or AGN variability phases.
    \item \textbf{Periodic light curves:} Periodic light curves with stable periods and distinctive Fourier amplitude‑ and phase‑ratios flag pulsators and eclipsing binaries rather than one‑off transients.
    \item \textbf{Secondary maxima:} Filter‑specific secondary maxima or shoulders in red/near‑IR bands—prominent in SNeIa—are morphological features absent in most core‑collapse SNe.
    \item \textbf{Monotonic flux trends:} Locally smooth, monotonic flux trends across one or multiple bands (plateaus, linear decays) capture physical evolution stages and help distinguish SNII‑P, SNII‑L, and related classes.
\end{enumerate}


%% file: figures/prompts/fig_supernova_explanation_prompt.tex
\begin{figure*}[ht]
  \centering
  \begin{promptbox}
What is the astrophysical classification of the following time series? Here are the possible labels you can use: RR-Lyrae (RRL), peculiar type Ia supernova (SNIa-91bg), type Ia supernova (SNIa), superluminous supernova (SLSN-I), type II supernova (SNII), microlens-single (mu-Lens-Single), eclipsing binary (EB), M-dwarf, kilonova (KN), tidal disruption event (TDE), peculiar type Ia supernova (SNIax), type Ibc supernova (SNIbc), Mira variable, and active galactic nuclei (AGN).

Each input is a multivariate time series visualized as a scatter plot image. The x-axis represents time, and the y-axis represents the flux measurement value. Each point corresponds to an observation at a specific timestamp and wavelength. Different wavelengths are color-coded, and observational uncertainty is shown using vertical error bars.

Even if the classification is uncertain or ambiguous, select the most likely label based on the observed visual patterns and provide a brief explanation that justifies your choice.

[BASELINE_PROMPT]

Your response should be 2 lines, formatted as follows:
Label: <astrophysical classification label>
Explanation: <explanation, as described above>

Here is the time series data for you to classify.
  \end{promptbox}
\caption{Supernova Explanation Prompt}
\label{fig:supernova_prompt}
\end{figure*}

%% file: sections/datasets/politeness.tex
\subsection{Politeness}
\textbf{Task.} Understanding how linguistic styles, like politeness, vary across cultures is necessary for building better communication, translation, and conversation-focused systems. \cite{holmes2012politeness, havaldar2023multilingual}. Today's LLMs exhibit large amounts of cultural bias \cite{havaldar2024building}, and understanding nuances in cultural differences can help encourage cultural adaptation in models. We use the holistic politeness dataset from \citet{havaldar2025comparingstyleslanguagescrosscultural}, which consists of conversational utterances between editors from Wikipedia talk pages, annotated by native speakers from four distinct cultures.

\textbf{Data Selection \& Preprocessing.} We sample 100 examples from the data, balanced equally across classes (rude, slightly rude, neutral, slightly polite, polite) and languages (English, Spanish, Japanese, Chinese). 

\input{figures/prompts/fig_politeness_explanation_prompt}

\textbf{Explanation Prompt.} We show the prompt in \cref{fig:politeness_prompt}. We replace \texttt{``[BASELINE\_PROMPT]''} with one of four prompting strategies shown in \cref{fig:baseline_prompts}.

\textbf{Expert Criteria.} The expert-validated criteria for expert alignment calculation are listed below:

\begin{enumerate}
\small
\itemsep 0em
    \item \textbf{Honorifics and Formal Address:} The presence of respectful or formal address forms (e.g., “sir,” “usted,”) signals politeness by expressing deference to the hearer’s status or social distance.
    \item \textbf{Courteous Politeness Markers:} Words such as “please,” “kindly,” or their multilingual variants soften requests and reflect courteous intent.
    \item \textbf{Gratitude Expressions:} Use of expressions like “thank you,” “thanks,” or “I appreciate it” signals recognition of the other’s contribution and positive face.
    \item \textbf{Apologies and Acknowledgment of Fault:} Phrases such as “sorry” or “I apologize” express humility and repair social breaches, marking a clear politeness strategy.
    \item \textbf{Indirect and Modal Requests:} Requests using modal verbs (“could you,” “would you”) or softening cues like “by the way” reduce imposition and signal respect for the hearer’s autonomy.
    \item \textbf{Hedging and Tentative Language:} Words like “I think,” “maybe,” or “usually” lower assertion strength and make statements more negotiable, reflecting interpersonal sensitivity.
    \item \textbf{Inclusive Pronouns and Group-Oriented Phrasing:} Use of “we,” “our,” or “together” expresses solidarity and reduces hierarchical distance in requests or critiques.
    \item \textbf{Greeting and Interaction Initiation:} Opening with a salutation (“hi,” “hello”) creates a cooperative tone and frames the conversation positively.
    \item \textbf{Compliments and Praise:} Positive evaluations (“great,” “awesome,” “neat”) attend to the hearer’s positive face and foster a friendly environment.
    \item \textbf{Softened Disagreement or Face-Saving Critique:} When disagreeing, the use of softeners, partial agreements, or concern for clarity preserves the hearer’s dignity.
    \item \textbf{Urgency or Immediacy of Language:} Utterances emphasizing emergency or speed (“asap,” “immediately”) can heighten perceived imposition and reduce politeness if not softened.
    \item \textbf{Avoidance of Profanity or Negative Emotion:} The presence of strong negative words or swearing is a key indicator of rudeness and face threat.
    \item \textbf{Bluntness and Direct Commands:} Requests lacking modal verbs or mitigation (“Do this”) are perceived as less polite due to their imperative structure.
    \item \textbf{Empathy or Emotional Support:} Recognizing the hearer’s emotional context or challenges is a politeness strategy of concern and goodwill.
    \item \textbf{First-Person Subjectivity Markers:} Statements that begin with “I think,” “I feel,” or “In my view” convey humility and subjectivity, reducing imposition.
    \item \textbf{Second Person Responsibility or Engagement:} Sentences starting with “you” or directly addressing the hearer can either signal engagement or come across as accusatory, depending on context and tone.
    \item \textbf{Questions as Indirect Strategies:} Questions (“what do you think?” or “could you clarify?”) reduce imposition by inviting rather than demanding input.
    \item \textbf{Discourse Management with Markers:} Use of discourse markers like “so,” “then,” “but” organizes conversation flow and may help manage face needs in conflict or negotiation.
    \item \textbf{Ingroup Language and Informality:} Use of group-identifying slang or casual expressions (“mate,” “dude,” “bro”) may foster solidarity or seem disrespectful, depending on relational norms.
\end{enumerate}


%% file: figures/prompts/fig_politeness_explanation_prompt.tex
\begin{figure*}[ht]
  \centering
  \begin{promptbox}
What is the politeness of the following utterance on a scale of 1-5? Use the following scale:
1: extremely rude
2: somewhat rude
3: neutral
4: somewhat polite
5: extremely polite

[BASELINE_PROMPT]

Your response should be 2 lines, formatted as follows:
Rating: <politeness rating>
Explanation: <explanation, as described above>

Utterance:
  \end{promptbox}
\caption{Politeness Explanation Prompt}
\label{fig:politeness_prompt}
\end{figure*}

%% file: sections/datasets/emotion.tex
\subsection{Emotion}

\textbf{Task.} Understanding and classifying emotion is important for tasks like therapy, mental health diagnoses, etc. \cite{denzin1984understanding}. Emotion is often expressed implicitly, and understanding such cues can also aid in building LLM systems that handle implied language understanding well \cite{havaldar2025entailed}. We use the GoEmotions dataset from \citet{demszky2020goemotions}, consisting of Reddit comments that have been human-annotated for one of 27 emotions (or neutral, if no emotion is present).

\textbf{Data Selection \& Preprocessing.} We sample 100 examples from the data, balanced equally across 28 emotion classes, including neutral. We additionally ensure the comment is over 20 characters, to remove noisy data points and ensure each comment contains enough information for the LLM to make an accurate classification. 

\input{figures/prompts/fig_emotion_explanation_prompt}

\textbf{Explanation Prompt.} We show the prompt in \cref{fig:emotion_prompt}. We replace \texttt{``[BASELINE\_PROMPT]''} with one of four prompting strategies shown in \cref{fig:baseline_prompts}.

\textbf{Expert Criteria.} The expert-validated criteria for expert alignment calculation are listed below:
\begin{enumerate}
\small
\itemsep 0em
    \item \textbf{Valence:} Decide if the overall tone is pleasant or unpleasant; positive tones suggest joy or admiration, negative tones suggest sadness or anger.
    \item \textbf{Arousal:} Gauge how energized the wording is—calm phrasing implies low arousal emotions, intense phrasing implies high arousal emotions.
    \item \textbf{Emotion Words \& Emojis:} Look for direct emotion terms or emoticons that explicitly name the feeling.
    \item \textbf{Expressive Punctuation:} Multiple exclamation marks, ALL‑CAPS, or stretched spellings signal higher emotional intensity.
    \item \textbf{Humor/Laughter Markers:} Tokens like “haha,” “lol,” or laughing emojis reliably indicate amusement.
    \item \textbf{Confusion Phrases:} Statements such as “I don’t get it” clearly mark confusion.
    \item \textbf{Curiosity Questions:} Genuine information‑seeking phrases (“I wonder…”, “why is…?”) point to curiosity.
    \item \textbf{Surprise Exclamations:} Reactions of astonishment (“No way!”, “I can’t believe it!”) denote surprise.
    \item \textbf{Threat/Worry Language:} References to danger or fear (“I’m scared,” “terrifying”) signal fear or nervousness.
    \item \textbf{Loss or Let‑Down Words:} Mentions of loss or disappointment cue sadness, disappointment, or grief.
    \item \textbf{Other‑Blame Statements:} Assigning fault to someone else for a bad outcome suggests anger or disapproval.
    \item \textbf{Self‑Blame \& Apologies:} Admitting fault and saying “I’m sorry” marks remorse.
    \item \textbf{Aversion Terms:} Words like “gross,” “nasty,” or “disgusting” point to disgust.
    \item \textbf{Praise \& Compliments:} Positive evaluations of someone’s actions show admiration or approval.
    \item \textbf{Gratitude Expressions:} Phrases such as “thanks” or “much appreciated” indicate gratitude.
    \item \textbf{Affection \& Care Words:} Loving or nurturing language (“love this,” “sending hugs”) signals love or caring.
    \item \textbf{Self‑Credit Statements:} Boasting about one’s own success (“I nailed it”) signals pride.
    \item \textbf{Relief Indicators:} Release phrases like “phew,” “finally over,” or “what a relief” mark relief after stress ends.
\end{enumerate}

%% file: figures/prompts/fig_emotion_explanation_prompt.tex
\begin{figure*}[ht]
  \centering
  \begin{promptbox}
What is the emotion of the following text? Here are the possible labels you could use: admiration, amusement, anger, annoyance, approval, caring, confusion, curiosity, desire, disappointment, disapproval, disgust, embarrassment, excitement, fear, gratitude, grief, joy, love, nervousness, optimism, pride, realization, relief, remorse, sadness, surprise, or neutral.

[BASELINE_PROMPT]

Your response should be 2 lines, formatted as follows:
Label: <emotion label>
Explanation: <explanation, as described above>

Here is the text for you to classify. Please ensure the emotion label is in the given list.
Text: 
  \end{promptbox}
\caption{Emotion Explanation Prompt}
\label{fig:emotion_prompt}
\end{figure*}

%% file: sections/datasets/cholec.tex
\subsection{Laparoscopic Cholecystectomy}

\textbf{Task.}
The task is to identify the safe and unsafe regions for incision.
We used the open-source subset of data from~\citep{madani2022artificial}, which consists of surgeon-annotated images taken from video frames from the 
M2CAI16 workflow challenge~\citep{stauder2016tum} and Cholec80~\citep{twinanda2016endonet} datasets.
This consists of 1015 surgeon-annotated images.

\textbf{Data Selection \& Preprocessing.}
We selected the first 100 items from the test set where the safe and unsafe regions were of nontrivial area.
Each item has three components: an image of dimensions 640 pixels wide by 360 pixels high, a binary mask of the safe regions of the same dimensions, and a binary mask of the unsafe regions of the same dimensions.

To convert the task into a form easily solvable by the available APIs, our objective was to have the LLM output a small list of numbers that identify the safe and unsafe regions.
This is achieved by using square grids of size 40 to discretize each of the safe and unsafe masks, separating them into \(144 = (640/40) \times (360/40)\) disjoint regions.
One can then use an integer inclusively ranging from \(0\) to \(143\) to uniquely identify these patches.
The LLM was to then output two lists with numbers from this range: a ``safe list'' that denotes its prediction of the safe region, and an ``unsafe list'' predicting the unsafe region.

\textbf{Explanation Prompt.}
We show the prompt in~\cref{fig:cholec_explanation_prompt}. We replace \texttt{[BASELINE\_PROMPT]} with one of four prompting strategies shown in \cref{fig:baseline_prompts}.

\textbf{Expert Criteria.} The expert-validated criteria for expert alignment calculation are listed below:

\begin{enumerate}
\small
\itemsep 0em
    \item[1.] \textbf{Calot's triangle cleared:} Hepatocystic triangle must be fully cleared of fat/fibrosis so that its boundaries are unmistakable.
    \item[2.] \textbf{Cystic plate exposed:} The lower third of the gallbladder must be dissected off the liver to reveal the shiny cystic plate and ensure the correct dissection plane.
    \item[3.] \textbf{Only two structures visible:} Only the cystic duct and cystic artery should be seen entering the gallbladder before any clipping or cutting.
    \item[4.] \textbf{Above the R4U line:} Dissection must remain cephalad to an imaginary line from Rouviere's sulcus to liver segment IV umbilical fissure to avoid the common bile duct. Dissection should be carried out along the inferior edge of the gallbladder (well above the line of safety).
    \item[5.] \textbf{Infundibulum start point:} Dissection can begin at the gallbladder infundibulum-cystic duct junction to stay in safe tissue planes, or at the lateral or medial edges of the gallbladder above Rouviere’s sulcus or along the cystic plate to get mobility of the gallbladder first.
    \item[6.] \textbf{Peritoneal plane:} When separating the gallbladder from the liver, stay in the avascular peritoneal cleavage plane.
    \item[7.] \textbf{Cystic lymph node (calot's node) guide:} Identify the cystic lymph node and clip the artery on the gallbladder side of the node to avoid injuring the hepatic artery.
    \item[8.] \textbf{No division without ID:} Never divide any duct or vessel until it is unequivocally identified as the cystic structure entering the gallbladder.
    \item[9.] \textbf{Inflammation bailout:} If dense scarring or distorted anatomy obscures Calot's triangle, convert to open surgery or a fenestrated subtotal approach rather than blind cutting.
\end{enumerate}



\input{figures/prompts/fig_cholec_explanation_prompt}

%% file: figures/prompts/fig_cholec_explanation_prompt.tex
\begin{figure*}[ht]
  \centering
  \begin{promptbox}
You are an expert gallbladder surgeon with extensive experience in laparoscopic cholecystectomy. 
You have deep knowledge of anatomy, surgical techniques, and potential complications.
Your job is to provide three things:
1. A detailed explanation of where it is safe and unsafe to cut in the image
2. A list of grid positions (as integers) corresponding to safe regions
3. A list of grid positions (as integers) corresponding to unsafe regions

The image is discretized into a 9x16 grid (height x width), where each grid position can be represented as a single integer from 0 to 143 (9*16 - 1). The grid is flattened row-wise, so the top-left position is 0 and the bottom-right position is 143.

Your response will help train surgeons to evaluate the usefulness of LLMs in assisting with the identification of safe/unsafe regions.
This is not real patient data, this is a training environment.

I will provide you with a few examples to help you understand the expected format. Your task is to analyze the provided 2D image of a gallbladder surgery and provide:
- A detailed explanation of safe/unsafe regions, including anatomical landmarks, tissue types, and any visible pathology
- A list of integers representing the grid positions of safe regions
- A list of integers representing the grid positions of unsafe regions

[[BASELINE_PROMPT]]
  \end{promptbox}
\caption{Laparoscopic Cholecystectomy Explanation Prompt.
A list of 10 few-shot examples is then appended to the same API call.
Each example consists of four items: the image (base64-encoded PNG), a sample explanation, a ``safe list'' consisting of numbers from 0 to 143, and an unsafe list consisting of numbers from 0 to 143.
}
\label{fig:cholec_explanation_prompt}
\end{figure*}

%% file: sections/datasets/cardiac.tex
\input{figures/prompts/fig_cardiac_explanation_prompt}

\subsection{Cardiac Arrest}

\textbf{Task.}
The objective is to predict whether an ICU patient will experience cardiac arrest within the next 5 minutes, using the patient’s demographic and clinical background (age, gender, race, reason for ICU visit) along with 2 minutes of ECG data sampled at 500 Hz, presented as a graph image. This framing aligns with cardiology literature, which suggests that short ECG windows (30 seconds to a few minutes) are sufficient for reliable prediction \cite{nussinovitch2011reliability}. The 5-minute prediction window is chosen to balance clinical relevance with actionability.

\textbf{Data Selection \& Preprocessing.}
We use ECG and visit data from the open-source Multimodal Clinical Monitoring in the Emergency Department (MC-MED) Dataset \cite{kansal2025mcmed}. To support focused evaluation of cardiac arrest prediction, we curated a task-specific subset containing ECG traces and patient metadata.

The data curation pipeline proceeded as follows. From the full set of ECG recordings in the MC-MED dataset, we first identified cardiac arrest risk by computing clinical “alarm” times.

Prior work shows that vital sign abnormalities are predictive of outcomes \cite{Candel2022-bo, chen2023multimodal}. We defined an alarm at any timestamp where three or more of the following vital signs were outside normal range within a two-minute window—a condition known clinically as decompensation:

\begin{itemize}
\vspace{-0.2cm}
\itemsep 0em
    \item Heart rate (HR): < 40 or > 130 bpm 
    \item Respiratory rate (RR): < 8 or > 30 breaths/min
    \item Oxygen saturation (SpO2): < 90\% 
    \item Mean arterial pressure (MAP): < 65 or > 120 mmHg
\end{itemize}

Each example was labeled 'Yes' if an alarm was present, and 'No' otherwise. For positive cases, we sampled a random cutoff time 1–300 seconds before the alarm and extracted the preceding 2 minutes of ECG data. For negative cases, we used the first 2 minutes of ECG data. We also added patient metadata—age, gender, race, and ICU admission reason—using information from the MC-MED visit records. To ensure diversity, each example came from a unique patient; for positives, we only used the visit containing the alarm.

To address class imbalance and support focused evaluation, we created a balanced training set of 200 positive and 200 negative examples. The validation and test sets each contain 50 examples.

\textbf{Explanation Prompt.}
Figure~\ref{fig:cardiac_prompt} shows the prompt used to generate explanations for predicting whether an ICU patient will experience cardiac arrest within 5 minutes, based on 2 minutes of ECG data along with age, gender, race, and ICU admission reason. We replace \texttt{[BASELINE\_PROMPT]} with one of four prompting strategies shown in \cref{fig:baseline_prompts}. The ECG is provided as a graph image of p-signal values sampled at 500 Hz over a 2-minute window, with labeled axes. While we considered supplying the raw signal as text, the input token limits of current LLMs made this infeasible.

\textbf{Expert Criteria.} The expert-validated criteria for expert alignment calculation are listed below:
\begin{enumerate}
\small
\itemsep 0em
  \item \textbf{Ventricular Ectopy / NSVT}: Runs of non-sustained ventricular tachycardia (NSVT) or frequent premature ventricular contractions may indicate electrical instability in critically ill patients, particularly those with underlying coronary disease or cardiomyopathy. While NSVT is generally considered a more benign form of ventricular tachyarrhythmia, its presence reflects transient abnormal rhythms originating from the lower chambers and does not necessarily always signal impending cardiac arrest.

  \item \textbf{Bradycardia or Heart-Rate Drop}: The onset of significant bradycardia or a sudden $\geq 30\%$ decline in heart rate is a well-documented precursor to in-hospital cardiac arrest (often preceding pulseless electrical activity or asystole) and should be treated as an alarm sign.

  \item \textbf{QRS Widening (Conduction Delay)}: New or progressive prolongation of the QRS duration on the ECG is an ominous finding in the ICU, often observed in the minutes before cardiac arrest and associated with higher mortality due to deteriorating ventricular conduction.

  \item \textbf{Dynamic ST-Segment Changes}: Acute ischemic changes on continuous ECG (notably ST-segment elevation or depression) signal low blood flow in the coronary arteries, indicating myocardial infarction or injury, and may precede imminent ventricular fibrillation or cardiac arrest. As blood flow continues to decrease, the ischemic heart can fibrillate or go into VT/VF (i.e., cardiac arrest). Although ST segment changes are common, the link to cardiac arrest is rare but possible.

  \item \textbf{QTc Prolongation}: Prolongation of the corrected QT (QTc) interval reflects abnormal ventricular repolarization and increases vulnerability to malignant arrhythmias such as torsades de pointes. QTc prolongation may result from medications, electrolyte disturbances (e.g., hypokalemia or hypomagnesemia), or underlying cardiac disease and is a known precursor to sudden cardiac arrest.

  \item \textbf{Severe Hyperkalemia Signs}: Electrocardiographic signs of severe hyperkalemia (such as peaked T-waves, loss of P-waves, and a widening QRS complex) herald an impending arrest -- as potassium levels rise, the ECG may evolve to a sine-wave pattern and typically culminate in ventricular fibrillation or asystole without immediate intervention. Hyperkalemia is a frequent cause of in-hospital cardiac arrest especially among patients on dialysis / end stage renal disease. Looking for signs of hyperkalemia can be important to understand risk of cardiac arrest, especially in selected populations.

  \item \textbf{Advanced Age}: Increasing age is a major risk factor for cardiac arrest, with events being uncommon in younger patients and substantially more frequent in older ICU populations. Patients over 65 years are more prone to sudden deterioration due to reduced physiologic reserve and a higher burden of cardiovascular disease.

  \item \textbf{Male Sex}: Male gender is associated with a higher incidence of cardiac arrest, as most cardiac arrests occur in men (with women’s risk rising post-menopause).

  \item \textbf{Underlying Cardiac Disease}: The presence of serious cardiac conditions -- such as coronary artery disease (especially a recent myocardial infarction) or severe heart failure -- greatly elevates short-term cardiac arrest risk by creating an electrically and hemodynamically unstable myocardium.

  \item \textbf{Critical Illness (Sepsis/Shock)}: Severe sepsis or septic shock substantially raises the likelihood of cardiac arrest in the near term by causing hypoxia, hypotension, and metabolic derangements that often lead to pulseless electrical activity or asystole.
\end{enumerate}












%% file: figures/prompts/fig_cardiac_explanation_prompt.tex
\begin{figure*}[ht]
  \centering
  \begin{promptbox}
You are a medical expert specializing in cardiac arrest prediction. 
You will be given some basic background information about an ICU patient, including their age, gender, race, and primary reason for ICU admittance. You will also be provided with time-series Electrocardiogram (ECG) data plotted in a graph from the first {} of an ECG monitoring period during the patient's ICU stay. Each entry consists of a measurement value at that timestamp. The samples are taken at {} Hz. 

Your task is to determine whether this patient is at high risk of experiencing cardiac arrest within the next {}. Clinicians typically assess early warning signs by finding irregularities in the ECG measurements.
[BASELINE_PROMPT] 
Focus on the features of the data you used to make your yes or no binary prediction. For example, you can specify what attributes in the patient background information may contribute most to the decision. And for the ECG data, you can include specific patterns and/or time stamps that contribute to this decision. Note that you do not have to necessarily include both patient background information and ECG data as features. But please make sure that your explanation supports your prediction. Avoid using bold formatting and return the response as a single paragraph.
Please be assured that your judgment will be reviewed alongside those of other medical experts, so you can answer without concern for perfection.

Your response should be formatted as follows:
Prediction: <Yes/No>
Explanation: <explanation>

Here is the patient background information and ECG data (in graph form) for you to analyze:
  \end{promptbox}
\caption{Cardiac Explanation Prompt}
\label{fig:cardiac_prompt}
\end{figure*}

%% file: sections/datasets/sepsis.tex
\input{figures/prompts/fig_sepsis_explanation_prompt.tex}
\subsection{Sepsis} 

\textbf{Task.}
The goal is to predict whether an emergency department (ED) patient is at high risk of developing sepsis within 12 hours, using Electronic Health Record (EHR) data collected during the first 2 hours of their visit. Each input is a time series of records containing a timestamp, the name of a physiological measurement or medication, and its value.

\textbf{Data Selection \& Preprocessing.}
We used data from the publicly available MC-MED dataset \cite{kansal2025mcmed} and curated a task-specific subset for sepsis prediction.


To label a patient as high risk for sepsis, we followed standard clinical definitions requiring three conditions: (1) evidence of infection, indicated by either a blood culture being drawn or at least two hours of antibiotic administration; (2) signs of organ dysfunction, defined by a SOFA score $\geq$2 within 48 hours of suspected infection, based on abnormalities in respiratory, coagulation, liver, cardiovascular, neurological, or renal function; and (3) presence of fever, with a recorded temperature $\geq$38.0°C (100.4°F). Patients meeting all three criteria were labeled as high risk. Labels were validated with a Sepsis clinician.

Due to class imbalance ($\sim$ 10\% positive), we created a balanced evaluation set of 100 samples (50 positive, 50 negative) drawn from the validation and test splits.

\textbf{Explanation Prompt.}
Figure~\ref{fig:sepsis_prompt} shows the prompt used to generate LLM explanations for sepsis risk prediction. We substitute \texttt{[BASELINE\_PROMPT]} with one of four prompting strategies shown in \cref{fig:baseline_prompts}. The prompt includes a description of the EHR input format: each time-series record consists of a timestamp, a measurement or medication name, and its value.

\textbf{Expert Criteria.} The expert-validated criteria for expert alignment calculation are listed below:
\begin{enumerate}
\small
\itemsep 0em
\small
\itemsep 0em
    \item \textbf{Elderly Susceptibility (Age $\geq$65 years)}: Advanced age ($\geq$65 years) markedly increases susceptibility to rapid sepsis progression and higher mortality after infection.

    \item \textbf{SIRS Positivity ($\geq$2 Criteria)}: Presence of $\geq$2 SIRS criteria—temperature $>$38$^\circ$C or $<$36$^\circ$C, heart rate $>$90 bpm, respiratory rate $>$20/min or PaCO$_2$ $<$32 mmHg, or WBC $>$12,000/$\mu$L or $<$4,000/$\mu$L—identifies systemic inflammation consistent with early sepsis.

    \item \textbf{High qSOFA Score ($\geq$2)}: A qSOFA score $\geq$2 (respiratory rate $\geq$22/min, systolic BP $\leq$100 mmHg, or altered mentation) flags high risk of sepsis-related organ dysfunction and mortality.

    \item \textbf{Elevated NEWS Score ($\geq$5 points)}: A National Early Warning Score (NEWS) of $\geq$5–7 derived from deranged vitals predicts imminent clinical deterioration compatible with sepsis.

    \item \textbf{Elevated Serum Lactate ($\geq$2 mmol/L)}: Serum lactate $\geq$2 mmol/L within the first 2 hours signals tissue hypoperfusion and markedly elevates sepsis mortality risk.

    \item \textbf{Elevated Shock Index ($\geq$1.0)}: Shock index (heart rate $\div$ systolic BP) $\geq$1.0—or a rise $\geq$0.3 from baseline—denotes haemodynamic instability and a high probability of severe sepsis.

    \item \textbf{Sepsis-Associated Hypotension (SBP $<$90 mmHg or MAP $<$70 mmHg, or $\geq$40 mmHg drop)}: Sepsis-associated hypotension, defined as SBP $<$90 mmHg, MAP $<$70 mmHg, or a $\geq$40 mmHg drop from baseline, indicates progression toward septic shock.

    \item \textbf{SOFA Score Increase ($\geq$2 points)}: An increase of $\geq$2 points in any SOFA component—e.g., PaO$_2$/FiO$_2$ $<$300, platelets $<$100$\times$10$^9$/L, bilirubin $>$2 mg/dL, creatinine $>$2 mg/dL, or GCS $<$12—confirms new organ dysfunction and high sepsis risk.

    \item \textbf{Early Antibiotic/Culture Orders (within 2 hours)}: Administration of broad-spectrum antibiotics or drawing of blood cultures within the first 2 hours signifies clinician suspicion of serious infection and should anchor sepsis risk assessment.
\end{enumerate}

%% file: figures/prompts/fig_sepsis_explanation_prompt.tex
\begin{figure*}[ht]
  \centering
  \begin{promptbox}
What is the sepsis risk prediction for the following time series? Here are the possible labels you can use: Yes (the patient is at high risk of developing sepsis within 12 hours) or No (the patient is not at high risk of developing sepsis within 12 hours).
The time series consists of Electronic Health Record (EHR) data collected during the first 2 hours of the patient's emergency department (ED) admission. Each entry includes a timestamp, the name of a measurement or medication, and its corresponding value.

[BASELINE_PROMPT]

Your response should be 2 lines, formatted as follows:
Label: <prediction label>
Explanation: <explanation, as described above>

Here is the text for you to classify.
  \end{promptbox}
\caption{Sepsis Explanation Prompt}
\label{fig:sepsis_prompt}
\end{figure*}